\documentclass[12pt]{article}

\usepackage{amsmath,amsfonts,bm,dsfont,amsthm}

\def\eqref#1{equation~\ref{#1}}

\def\1{\bm{1}}

\DeclareMathAlphabet{\mathsfit}{\encodingdefault}{\sfdefault}{m}{sl}
\SetMathAlphabet{\mathsfit}{bold}{\encodingdefault}{\sfdefault}{bx}{n}

\DeclareMathOperator{\Tr}{Tr}

\renewcommand{\min}[1]{\underset{#1}{\text{min}}\,}

\newcommand{\iid}[0]{\overset{iid}{\sim}}
\newcommand{\argmin}[1]{\underset{#1}{\text{arg min}}\,}
\newcommand{\argmax}[1]{\underset{#1}{\text{arg max}}\,}

\newcommand{\expectation}[1]{\underset{#1}{\mathbb{E}}}

\newcommand{\pr}[0]{\text{Pr}}

\newtheorem{remark}{Remark}[section]

\usepackage{graphicx}
\usepackage[acronym]{glossaries}

\usepackage{algpseudocode}
\usepackage{colortbl}
\usepackage{nicefrac}
\usepackage{subcaption}
\usepackage{booktabs}
\usepackage{multirow}
\usepackage{url}
\usepackage{fontawesome}
\usepackage{xcolor, colortbl}
\usepackage{tcolorbox}

\definecolor{Gray}{gray}{0.85}
\newcolumntype{a}{>{\columncolor{Gray}}c}

\definecolor{pastelgreen}{RGB}{102, 204, 102}
\definecolor{pastelred}{RGB}{255, 102, 102}

\usepackage[linesnumbered,ruled,vlined]{algorithm2e}

\SetKwComment{Comment}{\color{green!50!black}// }{}

\SetKwProg{Function}{function}{}{}

\renewcommand{\min}[1]{\underset{#1}{\text{min}}\,}

\usepackage{amsmath, amsfonts}

\newacronym{ot}{OT}{Optimal Transport}
\newacronym{da}{DA}{Domain Adaptation}
\newacronym{uda}{UDA}{Unsupervised Domain Adaptation}
\newacronym{gmm}{GMM}{Gaussian Mixture Model}
\newacronym{em}{EM}{Expectation Maximization}
\newacronym{gmmot}{GMMOT}{GMM OT}
\newacronym{otda}{OTDA}{OT for DA}
\newacronym{map}{MAP}{Maximum A Posteriori}
\newacronym{erm}{ERM}{Empirical Risk Minimization}
\newacronym{laot}{LaOT}{Linearly Alignable Optimal Transport}
\newacronym{cdfd}{CDFD}{Cross-Domain Fault Diagnosis}

\newacronym{cwru}{CWRU}{Case Western Reserve University}
\newacronym{cstr}{CSTR}{Continuous Stirred Tank Reactor}
\newacronym{te}{TE}{Tennessee Eastman}
\newacronym{pid}{PID}{Proportional, Integral, Derivative}
\newacronym{sne}{SNE}{Stochastic Neighbor Embeddings}

\newacronym{resnet}{ResNet}{Residual Networks}

\newacronym{dann}{DANN}{Domain Adversarial Neural Network}
\newacronym{dan}{DAN}{Domain Adaptive Network}
\newacronym{wdgrl}{WDGRL}{Wasserstein Distance Guided Representation Learning}
\newacronym{deepjdot}{DeepJDOT}{Deep Joint Distribution Optimal Transport}

\usepackage[left=2.5cm,right=2.5cm,top=2.5cm,bottom=2.5cm]{geometry}
\begin{document}


\begin{center}
{\Large
	{\sc  Optimal Transport for Domain Adaptation through Gaussian Mixture Models}
}
\bigskip

 Eduardo Fernandes Montesuma \& Fred Maurice Ngolè Mboula \& Antoine Souloumiac
\bigskip

{\it
$^{1}$ Université Paris-Saclay, CEA, List, F-91120 Palaiseau France\\
edumfern@gmail.com
}
\end{center}
\bigskip


{\bf Abstract.} Machine learning systems operate under the assumption that training and test data are sampled from a fixed probability distribution. However, this assumptions is rarely verified in practice, as the conditions upon which data was acquired are likely to change. In this context, the adaptation of the unsupervised domain requires minimal access to the data of the new conditions for learning models robust to changes in the data distribution. Optimal transport is a theoretically grounded tool for analyzing changes in distribution, especially as it allows the mapping between domains. However, these methods are usually computationally expensive as their complexity scales cubically with the number of samples. In this work, we explore optimal transport between Gaussian Mixture Models (GMMs), which is conveniently written in terms of the components of source and target GMMs. We experiment with 9 benchmarks, with a total of $85$ adaptation tasks, showing that our methods are more efficient than previous shallow domain adaptation methods, and they scale well with number of samples $n$ and dimensions $d$.


\begin{tcolorbox}

This is a pre-print. This paper was accepted at TMLR, see \url{https://openreview.net/forum?id=DCAeXwLenB}. Our code is available at,

\begin{center}
\faGithub\,\,\,\url{https://github.com/eddardd/gmm-otda/}
\end{center}
\end{tcolorbox}

\bigskip\bigskip

\section{Introduction}\label{sec:intro}

Supervised machine learning models are trained with significant amounts of labeled data, constituting a training set. The theory of generalization~\cite{redko2019advances}, provides a theoretical background that guarantees accurate predictions on unseen samples \emph{from the same distribution}. Nonetheless, these models are often forced to predict on related, but different data samples~\cite{quinonero2008dataset}. This distinction is modeled by a \emph{shift} in the probability distributions generating the data~\cite{sugiyama2007covariate}, which motivates the field of transfer learning~\cite{pan2009survey}, and more specifically the problem of \gls{uda}, in which models are adapted from a labeled \emph{source domain}, towards an unlabeled \emph{target domain}, following different distributions.

In this context, \gls{ot}~\cite{villani2009optimal,peyre2017computational} is a powerful, theoretically grounded framework for comparing and manipulating probability distributions~\cite{montesuma2023recent}. This framework works by computing a transportation strategy, that moves one probability distribution into the other at least effort. Based on this core idea, different methods have been proposed for \gls{uda} such as~\cite{courty2016otda,el2022hierarchical,chuang2023infoot} and~\cite{struckmeier2023learning}. 

However, this methodology faces a few challenges. For instance, \gls{ot} maps computed between discrete distributions are only defined for samples in the training set. Extrapolating these maps to new samples is the subject of intense research~\cite{perrot2016mapping,seguy2017large}. A possible workaround consists of using Gaussian approximations~\cite{flamary2019concentration,struckmeier2023learning}. While this approach effectively defines a mapping over the whole ambient space, its hypothesis do not reflect the possible sub-populations within the data, which are common in classification problems.

A natural solution to tackle multi-modality in data distributions is using \glspl{gmm}. A further advantage of this approach is considering the recently proposed \gls{gmmot} by~\cite{delon2020wasserstein}, which establishes an efficient, discrete problem between the components in the \glspl{gmm}. Furthermore, recent works have established the effectiveness of this idea for \emph{multi-source} domain adaptation~\cite{montesuma2024lighter}, notably through the use of mixture-Wasserstein barycenters.

Nonetheless, some questions on the use of \glspl{gmm} for \gls{uda} remain open. For instance,~\cite{delon2020wasserstein} propose different mapping strategies between \glspl{gmm}, but these either fail to map one \gls{gmm} into the other, or are subject to randomness when sampling transportation maps between \gls{gmm} components. Furthermore, if we label the components of \glspl{gmm}, it is natural to \emph{propagate} these labels, in the sense of~\cite{redko2019optimal} towards the target domain. This paper tackles these two questions.


\noindent\textbf{Summary of contributions.} In this paper, we propose 2 new strategies for \gls{uda} based on \glspl{gmm}. First, we use basic rules of probability theory for propagating the labels of source domain \gls{gmm} towards the target domain \gls{gmm}. We do so, by interpreting the \gls{gmmot} plan as the joint probability of source and target \gls{gmm} components. Second, we map samples from the source domain into the target domain based on the \gls{gmmot} plan. For a point in the source domain, our strategy consists of first estimating the component, in the source \gls{gmm}, most likely to have generated the sample. We then transport this point to components in the target domain, while assigning importance weights based on the \gls{gmmot} plan.


\noindent\textbf{Paper organization.} The rest of this paper is organized as follows. Section~\ref{sec:related_works} presents a few related works on \gls{ot} for \gls{uda}. Section~\ref{sec:preliminaries} covers the preliminaries on \gls{ot} and \gls{gmmot} theory. Section~\ref{sec:methodology} covers our methodological contributions. Section~\ref{sec:experiments} details our experiments and discussion on \gls{uda}. Finally, section~\ref{sec:conclusion} concludes this paper.

\noindent\textcolor{black}{\textbf{Notation.} We use uppercase letters $P$ and $Q$ to denote probability distributions, and $P_{S}$ and $P_{T}$ to denote source and target domain distributions. More generally, we use $\pr$ to denote probabilities. For instance, $\pr(Y=y|X=\mathbf{x})$ denotes the conditional probability of label $Y=y$ given a feature vector $X = \mathbf{x}$. Let $P$ be a distribution over feature vectors. We denote samples from $P$ as $\mathbf{x}^{(P)}$. We reserve $y^{(P)}$ for categorical labels (i.e., $1, \cdots, n_{c}$, for $n_{c}$ classes), and $\mathbf{y}^{(P)}$ for its one-hot encoding.}
\section{Related Works}\label{sec:related_works}

\textbf{Optimal transport for domain adaptation.} Optimal transport has been extensively employed for the design of algorithms~\cite{courty2016otda}, as well as analyzing the domain adaptation problem~\cite{redko2017theoretical}. The key idea of this method is to use the Kantorovich formulation to acquire a matching, known as transport plan, between source and target domain distributions. This matching defines a map between points in the source domain, towards the target domain, called barycentric mapping. Based on this idea, different methods have proposed improvements. For instance,~\cite{perrot2016mapping} proposed learning linear and kernelized extensions of the barycentric map through convex optimization.~\cite{el2022hierarchical} uses clustering for learning matching with additional structural dependencies.~\cite{flamary2019concentration} uses optimal transport between Gaussian distributions for estimating an affine mapping between source and target domains. More recently,~\cite{chuang2023infoot} proposed a method that leverages kernel density estimation for defining a new optimal transport problem based on information maximization.

\noindent\textbf{Gaussian-mixture based optimal transport.} An optimal transport problem involving Gaussian mixtures was initially proposed by~\cite{chen2018optimal}, in which a linear program between the components of the two mixtures is solved. This setting was further studied by~\cite{delon2020wasserstein}, who proved an interesting connection to a continuous optimal transport, when the transport plan is constrained to the set of Gaussian mixtures. Based on the framework of~\cite{delon2020wasserstein},~\cite{montesuma2024lighter} proposed the extension of multi-source domain adaptation algorithms of~\cite{montesuma2021icassp,montesuma2021cvpr} and~\cite{montesuma2023dadil}. However, these authors focused on performing adaptation through Wasserstein barycenters. Although based on the same framework, our work focuses on \emph{mapping samples and propagating labels} of Gaussian mixtures, especially for single-source domain adaptation.

\section{Theoretical Foundations}\label{sec:preliminaries}

\subsection{Optimal Transport}\label{sec:optimal_transport}

Founded by~\cite{monge1781memoire}, optimal transport is a field of mathematics concerned with transporting mass at least effort. Let $\mathcal{X}$ be a set and $\mathcal{P}(\mathcal{X})$ the set of probability distributions on $\mathcal{X}$. For $P, Q \in \mathcal{P}(\mathcal{X})$, the Monge formulation of the optimal transport problem is,
\begin{align}
    T^{\star} &= \underset{T:T_{\sharp}P=Q}{\text{arginf}} \int_{\mathcal{X}}c(x, T(x))dP(x),\label{eq:MongeEq}
\end{align}
where $T_{\sharp}P$ denotes the pushforward distribution~\cite[Problem 1.1]{santambrogio2015optimal} of $P$ by the map $T$, and $c:\mathcal{X}\times\mathcal{X}\rightarrow\mathbb{R}$, called the ground-cost, denotes the cost of sending $x$ to position $T(x)$.

Although equation~\ref{eq:MongeEq} is a formal description of the optimal transport problem, the constraint $T_{\sharp}P=Q$ poses technical difficulties. An alternative description was proposed by~\cite{kantorovich1942transfer}, in terms of an optimal transport \emph{plan} $\gamma$,
\begin{align}
    \textcolor{black}{\gamma^{\star} = \underset{\gamma \in \Gamma(P,Q)}{\text{arginf}}\int_{\mathcal{X}\times\mathcal{X}}c(x_{1},x_{2})d\gamma(x_{1},x_{2}),}\label{eq:KantorovichEq}
\end{align}
where $\Gamma(P, Q)$ is the set of joint distributions with marginals $P$ and $Q$. This formulation is simpler to analyze because the constraint $\gamma \in \Gamma(P, Q)$ is linear with respect to the optimization variable $\gamma$.

When $(\mathcal{X}, d)$ is a metric space, it is possible to define a distance on $\mathcal{P}(\mathcal{X})$ in terms of $d$. Let $\alpha \in [1,+\infty)$, and $c(\cdot,\cdot) = d(\cdot,\cdot)^{\alpha}$. One then has the so-called $\alpha-$Wasserstein distance:
\begin{align}
     \textcolor{black}{\mathcal{W}_{\alpha}(P, Q)^{\alpha}} &= \textcolor{black}{\underset{\gamma \in \Gamma(P,Q)}{\text{inf}}\int_{\mathcal{X}\times\mathcal{X}}c(x_{1},x_{2})d\gamma(x_{1},x_{2}).}\label{eq:WassersteinDist}
\end{align}
In the following, we do optimal transport on Euclidean spaces, i.e., $\mathcal{X} = \mathbb{R}^{d}$. In this case, it is natural to use $d(\mathbf{x}_{1},\mathbf{x}_{2}) = \lVert \mathbf{x}_{1} - \mathbf{x}_{2} \rVert_{2}$. Furthermore, we set $\alpha = 2$. Equations~\ref{eq:KantorovichEq} and~\ref{eq:WassersteinDist} are linear programs, where the optimization variable is the joint distribution $\gamma$. In the following, we discuss 3 particular cases where optimal transport either has a closed for, or is approximated by a finite problem, thus tractable by a computer.

\noindent\textbf{Empirical Case.} If we have samples $\{\mathbf{x}_{i}^{(P)}\}_{i=1}^{n}$ and $\{\mathbf{x}_{j}^{(Q)}\}_{j=1}^{m}$ with probabilities $p_{i}$ and $q_{j}$ respectively, we can make empirical approximations for $P$ and $Q$,
\begin{align}
    \hat{P}(\mathbf{x}) &= \sum_{i=1}^{n}p_{i}\delta(\mathbf{x}-\mathbf{x}_{i}^{(P)}).\label{eq:emp_distr}
\end{align}
The approximation in equation~\ref{eq:emp_distr} is at the core of discrete optimal transport~\cite{peyre2017computational}. If we plug equation~\ref{eq:emp_distr} into equation~\ref{eq:KantorovichEq}, the optimal transport problem becomes computable, i.e., it turns into a linear programming problem with $n \times m$ variables,
\begin{align}
    \gamma^{\star} &= \argmin{\gamma \in \Gamma(\hat{P}, \hat{Q})}\sum_{i=1}^{n}\sum_{j=1}^{m}\gamma_{ij}C_{ij},\label{eq:DiscreteKantorovich}
\end{align}
where $C_{ij} = c(\mathbf{x}_{i}^{(P)},\mathbf{x}_{j}^{(Q)})$. As a linear program, one should keep in mind that solving equation~\ref{eq:DiscreteKantorovich} has a complexity of $\mathcal{O}(n^{3}\log n)$. This complexity can be reduced by regularizing the problem in terms of the \gls{ot} plan entropy,
\begin{align}
    \gamma^{\star} &= \argmin{\gamma \in \Gamma(\hat{P}, \hat{Q})}\sum_{i=1}^{n}\sum_{j=1}^{m}\gamma_{ij}C_{ij}+\epsilon\sum_{i=1}^{n}\sum_{j=1}^{m}\gamma_{ij}(\log \gamma_{ij} - 1),\label{eq:Sinkhorn}
\end{align}
which can be solved through the celebrated Sinkhorn algorithm~\cite{cuturi2013sinkhorn}. A solution can be found with complexity $\mathcal{O}(Ln^{2})$, where $L$ is the number of iterations of the algorithm. \textcolor{black}{From the Kantorovich formulation, we can recover a correspondence between distributions through the barycentric map,}
\begin{align}
    T_{\gamma}(\mathbf{x}_{i}^{(P)}) &= \min{\mathbf{x} \in \mathcal{X}}\sum_{j=1}^{m}\gamma_{ij}c(\mathbf{x},\mathbf{x}_{j}^{(Q)}).\label{eq:BarycentricMapping}
\end{align}

\noindent\textbf{Gaussian Case.} When $P = \mathcal{N}(\mu^{(P)}, \Sigma^{(P)})$ (resp., $Q$), equation~\ref{eq:MongeEq} has a closed and affine form~\cite{takatsu2011wasserstein}, $T^{\star}(\mathbf{x}) = \mathbf{Ax}+\mathbf{b}$, where,
\begin{equation}
    \mathbf{A} = (\Sigma^{(P)})^{-\frac{1}{2}}((\Sigma^{(P)})^{\frac{1}{2}}\Sigma^{(Q)}(\Sigma^{(P)})^{\frac{1}{2}})^{\frac{1}{2}}(\Sigma^{(P)})^{-\frac{1}{2}}\text{, and }\mathbf{b} = \mu^{(Q)} - \mathbf{A}\mu^{(P)},\label{eq:gauss_map}
\end{equation}
and the Wasserstein distance takes the form,
\begin{equation}
    \mathcal{W}_{2}(P, Q)^{2} = \lVert \mu^{(P)} - \mu^{(Q)} \rVert^{2}_{2} + \Tr(\Sigma^{(P)} + \Sigma^{(Q)} + ((\Sigma^{(P)})^{\nicefrac{1}{2}}\Sigma^{(Q)}(\Sigma^{(P)})^{\nicefrac{1}{2}})^{\nicefrac{1}{2}}).\label{eq:gauss_w2}
\end{equation}
In contrast with empirical optimal transport, under the Gaussian hypothesis, computing the Wasserstein distance and a mapping between $P$ and $Q$ has sample-free complexity. Indeed, the complexity of equations~\ref{eq:gauss_map} and~\ref{eq:gauss_w2} is dominated by computing the square-root of the covariance matrix with complexity $\mathcal{O}(d^{3})$.

Furthermore, in high dimensions and when a only a few data points are available, estimating full covariance matrices is challenging. In these cases, it is useful to assume axis-aligned Gaussians, i.e., $\Sigma$ is a diagonal matrix with diagonal elements $\Sigma_{ii} = \sigma_{i}^{2}$. In this case, equations~\ref{eq:gauss_map} and~\ref{eq:gauss_w2} can be further simplified,
\begin{align}
    \mathbf{A} = \text{diag}(\nicefrac{\sigma^{(Q)}}{\sigma^{(P)}})\text{, }\mathbf{b} = \mu^{(Q)} - \mathbf{A}\mu^{(P)}\text{, and }\mathcal{W}_{2}(P, Q)^{2} = \lVert \mu^{(P)} - \mu^{(Q)} \rVert_{2}^{2} + \lVert \sigma^{(P)} - \sigma^{(Q)} \rVert_{2}^{2}.\label{eq:axis_aligned_gauss}
\end{align}

\begin{figure}[t]
    \centering
    \begin{subfigure}[b]{0.3\linewidth}
        \centering
        \includegraphics[width=\linewidth]{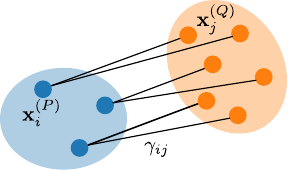}
        \caption{Empirical OT}
    \end{subfigure}\hfill
    \begin{subfigure}[b]{0.3\linewidth}
        \centering
        \includegraphics[width=\linewidth]{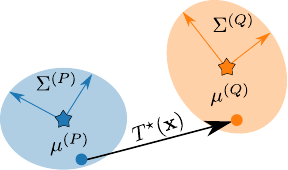}
        \caption{Gaussian OT}
    \end{subfigure}\hfill
    \begin{subfigure}[b]{0.3\linewidth}
        \centering
        \includegraphics[width=\linewidth]{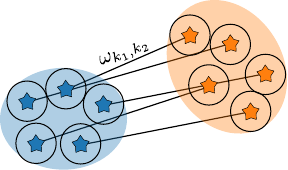}
        \caption{GMM-OT}
    \end{subfigure}
    \caption{\textbf{Comparison of different ways of solving \gls{ot}.} In empirical \gls{ot}, $P$ and $Q$ are approximated non-parametrically through their samples. In Gaussian \gls{ot}, $P$ and $Q$ are Gaussian distributions, and the mapping between these distributions is affine. In GMM-OT, $P$ and $Q$ are assumed to be \glspl{gmm}, and an \gls{ot} plan between components, $\omega$, defines an \gls{ot} plan between samples, $\gamma$.}
    \label{fig:ot_estimation}
\end{figure}

\noindent\textbf{Gaussian Mixture Case.} A Gaussian mixture corresponds to $P=\sum_{k=1}^{K}\pi_{k}^{(P)}P_{k}$, where $P_{k} = \mathcal{N}(\mu_{k}^{(P)},\Sigma_{k}^{(P)})$. As in~\cite{delon2020wasserstein}, we denote by $\text{GMM}_{d}(K)$ the set of distributions $P \in \mathcal{P}(\mathbb{R}^{d})$ written as a mixture of at most $K$ components. In this framework~\cite{delon2020wasserstein} explores the optimal transport problem~\ref{eq:KantorovichEq} under the constraint that $\gamma$ is a \gls{gmm} as well, i.e., $\gamma \in \Gamma(P,Q) \cap GMM_{2d}(\infty)$. This formulation is interesting because it is equivalent to a discrete and hierarchical problem~\cite[Proposition 4]{delon2020wasserstein} in terms of the \glspl{gmm}' components
\begin{align}
    \omega^{\star} = \text{GMMOT}(P, Q) = \argmin{\omega \in \Gamma(\pi^{(P)}, \pi^{(Q)})}\sum_{k_{1}=1}^{K_{P}}\sum_{k_{2}=1}^{K_{Q}}\omega_{k_{1},k_{2}}\mathcal{W}_{2}(P_{k_{1}},Q_{k_{2}})^{2}.\label{eq:GMMOT}
\end{align}
We call this problem \gls{gmmot}. This latter equation is an hierarchical optimal transport problem, i.e., a problem that involves itself an inner optimal transport. As in the previous cases, we have a notion of distance related to equation~\ref{eq:GMMOT},
\begin{align*}
    \mathcal{MW}_{2}(P, Q)^{2} = \sum_{k_{1}=1}^{K_{1}}\sum_{k_{2}=1}^{K_{2}}\omega_{k_{1},k_{2}}\mathcal{W}_{2}(P_{k_{1}},Q_{k_{2}})^{2}.
\end{align*}
We show an overview of different strategies for solving \gls{ot}, under different assumptions, in Figure~\ref{fig:ot_estimation}.

\begin{remark}{(Computational Complexity)}\label{remark:complexity}
    The overall complexity of the \gls{gmmot} in equation~\ref{eq:GMMOT} is $\mathcal{O}(K^{3}\log K)$. However, one should keep in mind that the ground-cost matrix must be computed beforehand, as given by equation~\ref{eq:gauss_w2}, which involves a complexity that scales with the dimension of the ambient space, due the matrix inversions and square-roots. The complexity of these operations is $\mathcal{O}(K^{2}d^{3})$ in general. However, assuming diagonal covariance matrices, the computational complexity is drastically reduced to $\mathcal{O}(K^{2}d)$, i.e., the complexity of computing $K^{2}$ Euclidean distances between $d-$dimensional vectors. \textcolor{black}{We refer readers to our appendix for a running time analysis of our method.}
\end{remark}

\begin{remark}{(Parameter Estimation)}
    Besides the computational advantage, using diagonal covariance matrices yields a simpler estimation problem for \glspl{gmm}. This is pivotal in \gls{da}, as the target domain likely does not have enough samples for the accurate estimation of complete covariance matrices. Furthermore, data is oftentimes high-dimensional, as feature spaces commonly involve thousands of features. This is a sharp contrast with previous works in \gls{gmmot}, such as~\cite{delon2020wasserstein} and~\cite{chen2018optimal}, which involved a few dimensions. \textcolor{black}{In our empirical validation (c.f., section~\ref{sec:visda}) we show that using diagonal covariances yields better adaptation performance in high dimensions.}
\end{remark}

\subsection{Learning Theory and Domain Adaptation}

In this paper, we deal with \gls{da} for classification. This latter problem can be formalized mathematically, as the learning of a function $h:\mathcal{X}\rightarrow\mathcal{Y}$, from a feature space $\mathcal{X}$ (e.g., $\mathbb{R}^{d}$) to a label space, $\mathcal{Y} = \{1,\cdots,n_{c}\}$ through samples of a probability distribution. As reviewed by~\cite{redko2019advances}, from the point of view of probability, there multiple ways of formalizing this problem. Here, we use the \gls{erm} framework of~\cite{vapnik2013nature}. For a probability distribution $P$, a loss function $\mathcal{L}:\mathcal{Y}\times\mathcal{Y} \rightarrow\mathbb{R}_{+}$, and a family of classifiers $\mathcal{H}$, one may define a notion of disagreement between pairs $h, h' \in \mathcal{H}$,
\begin{align}
    \mathcal{R}_{P}(h,h') = \expectation{\mathbf{x} \sim P}[\mathcal{L}(h(\mathbf{x}), h'(\mathbf{x}))].\label{eq:risk}
\end{align}
Equation~\ref{eq:risk} defines the risk of $h$ with respect $h'$. Given a ground-truth labeling function $h_{0}:\mathcal{X}\rightarrow\mathcal{Y}$, classification can be phrased in terms of minimizing the risk of $h$ with respect the ground-truth $h_{0}$, i.e.,
\begin{align*}
    h^{\star} = \argmin{h\in\mathcal{H}}\mathcal{R}_{P}(h,h_{0}).
\end{align*}
Henceforth, we adopt $\mathcal{R}_{P}(h) = \mathcal{R}_{P}(h, h_{0})$ in short. This formalization equates the problem of learning a classifier with an optimization problem. Nonetheless, in practice one does not have access to \emph{a priori} knowledge from $P$ nor $h_{0}$. In a more realistic scenario, one has samples $\{\mathbf{x}_{i}^{(P)}, y_{i}^{(P)}\}_{i=1}^{n}$, where $\mathbf{x}_{i}^{(P)} \iid P$, and $y_{i}^{(P)} = h_{0}(\mathbf{x}_{i}^{(P)})$. Based on these samples, one may estimate the risk empirically, by resorting to the approximation in equation~\ref{eq:emp_distr},
\begin{align}
    \hat{\mathcal{R}}_{P}(h) = \dfrac{1}{n}\sum_{i=1}^{n}\mathcal{L}(h(\mathbf{x}_{i}^{(P)}),y_{i}^{(P)})\text{, and }\hat{h} = \argmin{h\in\mathcal{H}}\hat{\mathcal{R}}_{P}(h).\label{eq:emp_risk}
\end{align}
In machine learning literature, the true risk $\mathcal{R}_{P}(h)$ is called \emph{generalization error}, i.e., the error that a classifier $h$ makes on samples from the distribution $P$. In contrast, algorithms usually minimize the \emph{training error}, $\hat{\mathcal{R}}_{P}(h)$, i.e., the error of $h$ on the particular examples available during training.


A key limitation of the presented theory is its assumption that data originates from a single probability distribution $P$. As discussed by~\cite{quinonero2008dataset}, this is seldom happens in practice. For instance, in fault diagnosis, process conditions influence the statistical properties of measured signals~\cite{montesuma2022cross}. As a result, generalization must be carried to a new, related probability distribution. This problem is known in the literature as \gls{da}, a sub-field within transfer learning.

As discussed by~\cite{pan2009survey}, in transfer learning, a domain is a pair $\mathcal{D} = (\mathcal{X}, P)$ of a feature space and a probability distribution over $\mathcal{X}$. Likewise, a task is a pair $\mathcal{T} = (\mathcal{Y},h_{0})$ of a label space and a ground-truth labeling function. Transfer learning is characterized by different source and target domains and tasks, i.e., $(\mathcal{D}_{S}, \mathcal{T}_{S}, \mathcal{D}_{T}, \mathcal{T}_{T})$, where at least one element from the source is different from the target. \textcolor{black}{In this work, we assume different domains $\mathcal{D}_{S} \neq \mathcal{D}_{T}$ but the same label space, $\mathcal{Y}_{S} = \mathcal{Y}_{T} = \{1,\cdots,n_{c}\}$}.

In this paper, we deal primarily with \emph{distributional shift}. In this case, we assume $\mathcal{X}_{S} = \mathcal{X}_{T} = \mathbb{R}^{d}$, so that source and target domains are characterized by different probability distributions $P_{S} \neq P_{T}$. Furthermore, we place ourselves in the \emph{unsupervised \gls{da}} setting, that is, we assume labeled source domain data, $\{\mathbf{x}_{i}^{(P_{S})}, y_{i}^{(P_{S})}\}_{i=1}^{n}$, and unlabeled target domain data $\{\mathbf{x}_{j}^{(P_{T})}\}_{j=1}^{m}$. Our goal is to use these samples to learn a classifier $h$ that works well on the target domain, i.e., that achieves small target risk $\mathcal{R}_{P_{T}}$.

\textcolor{black}{From the point of view of \gls{da}, the quality and similarity of source domain data plays a prominent role, since supervision comes from this distribution. For instance, if source domain data contains noisy labels, these can be transferred to the target domain, leading to poor results. Likewise, the success of domain adaptation is correlated with the distance, in distribution, between these two domains. We refer readers to~\cite{ben2010theory,redko2017theoretical} for further discussion.}

\noindent\textbf{Optimal Transport for Domain Adaptation} was proposed by~\cite{courty2016otda}, and primarily tries to match samples from $P$ to those of $Q$ based on the empirical \gls{ot} problem in equation~\ref{eq:DiscreteKantorovich}. After acquiring $\gamma^{\star}$, the authors propose mapping samples from $P$ towards those of $Q$ via the baryncetric map. This strategy effectively constitutes a new dataset $\{T_{\gamma^{\star}}(\mathbf{x}_{i}^{(P)}), y_{i}^{(P)}\}_{i=1}^{n}$, where $T_{\gamma^{\star}}$ is defined by equation~\ref{eq:BarycentricMapping}. Note, here, that under $T_{\gamma^{\star}}$, the source domain points \emph{carry their labels}, to the target domain. This operation is valid, as long as the conditionals $P_{S}(Y|X) = P_{T}(Y|T(X))$, which is restrictive, but reasonable under the covariate shift hypothesis. A few problems plague this strategy. First, $T_{\gamma^{\star}}$ is only defined on the support of $P$. The mapping of new points has been extensively studied in the literature~\cite{perrot2016mapping,seguy2017large,chuang2023infoot}. Second, it is desirable to have mappings with additional structure with respect the classes in \gls{da}. This problem is partially solved by considering special regularization schemes, as covered in~\cite{courty2016otda}. Third, this method is not scalable with respect the number of samples $n$, due its prohibitive complexity $\mathcal{O}(n^{3}\log n)$. In this paper, we offer a solution for the aforementioned problems through the \gls{gmm}-\gls{ot} framework of~\cite{delon2020wasserstein}. 
\section{Domain Adaptation via Optimal Transport between Gaussian Mixtures}\label{sec:methodology}

\begin{minipage}[t]{\textwidth}
\begin{algorithm}[H]
  \caption{Fitting procedure for GMMs.}
  \label{alg:em}
  \Function{\textcolor{black}{EM($\mathbf{X}^{(P)}, K_{P}$)}}{
    \For{$it=1,\cdots,n_{iter}$}{
        \Comment{Expectation Step}
        $\textcolor{black}{G_{i,k} = \dfrac{\pi_{k}^{(P)}\mathcal{N}(\mathbf{x}_{i}^{(P)}|\mu_{k}^{(P)},\Sigma_{k}^{(P)})}{\sum_{k'}\pi_{k'}^{(P)}\mathcal{N}(\mathbf{x}_{i}^{(P)}|\mu_{k'}^{(P)},\Sigma_{k'}^{(P)})}}$\;

        \Comment{Maximization Step}
        $\textcolor{black}{n_{k} = \sum_{i=1}^{n}G_{i,k}}$\;
        
        $\textcolor{black}{\mu^{(P)}_{k} \leftarrow \dfrac{1}{n_{k}}\sum_{i=1}^{n}G_{ik} \mathbf{x}_{i}^{(P)}}$\;

        $\textcolor{black}{\Sigma^{(P)}_{k} \leftarrow \dfrac{1}{n_{k}}\sum_{i=1}^{n}G_{i,k} (\mathbf{x}_{i}^{(P)} - \mu_{k}^{(P)})(\mathbf{x}_{i}^{(P)} - \mu_{k}^{(P)})^{T}}$\;

        $\textcolor{black}{\pi_{k}^{(P)} \leftarrow \dfrac{n_{k}}{n}}$\;
    }
    \Return{$\textcolor{black}{\mu^{(P)}, \Sigma^{(P)}, \pi^{(P)}}$}\;
  }
\end{algorithm}
\end{minipage}
\hfill
\begin{minipage}[t]{\textwidth}
\begin{algorithm}[H]
  \caption{Fitting procedure for labeled GMMs.}
  \label{alg:conditional_em}
  \Function{\textcolor{black}{ConditionalEM($\mathbf{X}^{(P)}, \mathbf{Y}^{(P)}, K_{P}$)}}{
    $\textcolor{black}{cpc \leftarrow K_{P} / n_{c}}$\;

    \For{$\textcolor{black}{y=1,\cdots,n_{c}}$}{
        \Comment{Samples from y-th class}
        $\textcolor{black}{\mathbf{X}^{(P_{y})} \leftarrow \{ \mathbf{x}_{i}^{(P)}: y_{i}^{(P)} = y \}}$\;

        \Comment{EM on conditionals}
        $\textcolor{black}{\mu^{(P_{y})}, \Sigma^{(P_{y})}, \pi^{(P_{y})} \leftarrow \text{EM}(\mathbf{X}^{(P_{y})}, cpc)}$\;

        $\textcolor{black}{\nu^{(P_{y})} \leftarrow \{\text{one\_hot}(y)\}_{k=1}^{cpc}}$\;
    }
    \Comment{Concatenates all parameters}
    $\textcolor{black}{\mu^{(P)} = \{\{ \mu_{k}^{(P_{y})} \}_{k=1}^{cpc}\}_{y=1}^{n_{c}}}$\;

    $\textcolor{black}{\nu^{(P)} = \{\{ \nu_{k}^{(P_{y})} \}_{k=1}^{cpc}\}_{y=1}^{n_{c}}}$\;

    $\textcolor{black}{\Sigma^{(P)} = \{\{ \Sigma_{k}^{(P_{y})} \}_{k=1}^{cpc}\}_{y=1}^{n_{c}}}$\;
    
    $\textcolor{black}{\pi^{(P)} = \biggr{\{}\biggr{\{} \dfrac{\pi^{(P_{y})}_{k}}{\sum_{k=1}^{cpc}\sum_{y=1}^{n_{c}}\pi_{k}^{(P_{y})} } \biggr{\}}_{k=1}^{cpc}\biggr{\}}_{y=1}^{n_{c}}}$\;
    
    \Return{$\textcolor{black}{\mu^{(P)}, \Sigma^{(P)}, \nu^{(P)}, \pi^{(P)}}$}\;
  }
\end{algorithm}
\end{minipage}

In this section, we develop new tools for \gls{da} through \gls{ot} between \glspl{gmm}. As we discussed in our preliminaries section, we are particularly interested in \gls{da} for classification. In this context, data is naturally multi-modal, which justifies the mixture modeling. \textcolor{black}{As we previously defined in Section~\ref{sec:preliminaries}, a \gls{gmm} is a mixture model with parameters $\{\mu_{k}^{(P)}, \Sigma_{k}^{(P)}, \pi_{k}^{(P)}\}_{k=1}^{K}$. These parameters can be determined through maximum likelihood:}
\begin{align}
    \textcolor{black}{\{\mu_{k}^{(P)}, \Sigma_{k}^{(P)}, \pi_{k}^{(P)}\}_{k=1}^{K} = \argmax{\{\mu_{k}, \Sigma_{k}, \pi_{k}\}_{k=1}^{K}} \sum_{i=1}^{n}\log P(\mathbf{x}_{i}^{(P)}),}\label{eq:em}
\end{align}
\textcolor{black}{where $P = \sum_{k=1}^{K}\pi_{k}^{K}P_{k}$. A practical approach for optimizing equation~\ref{eq:em} was proposed by~\cite{dempster1977maximum}, and is known as \gls{em}. We show a pseudo-code for this strategy in Algorithm~\ref{alg:em}. We refer readers to~\cite[Chapter 9]{bishop2006pattern} for further details on \glspl{gmm}.}

\textcolor{black}{In our approach, we need to define labels for the components of \glspl{gmm} in the transportation problem. We do so through an heuristic, that is, we model each $P_{y} = P(X|Y=y)$ through a \gls{gmm}, for $y=1,\cdots,n_{c}$. As a result, we fit a \gls{gmm} to the data $\mathbf{X}^{(P_{y})} = \{\mathbf{x}_{i}^{(P)}\}_{i:y_{i}^{(P)}=y}$, using $cpc = K_{P} / n_{c}$ components. Here, we conveniently choose $K_{P}$ as a multiple of $n_{c}$, for ensuring that $cpc$ is an integer. We present in Algorithm~\ref{alg:conditional_em} a pseudo-code for this strategy. Furthermore, we create a one-hot encoded vector of component labels, denoted $\nu^{(P)} \in (\Delta_{n_{c}})^{K}$, where $\nu_{k,y}^{(P)} = 1$ if the $k-$th component comes from the $y-$th class, and $0$ otherwise. We interpret the vector $\nu^{(P)}_{k}$ as the conditional probability $\pr(Y|K=k)$.}


In the following, we discuss two strategies for domain adaptation. The first, based on label propagation, leverages the optimal transport plan \emph{between components} to define pseudo-labels for the components of the target domain \gls{gmm}. The second, based on mapping estimation, leverages the hierarchical nature of the \gls{gmmot} problem for defining a map between source and target domain. \textcolor{black}{These methods are summarized in Algorithms~\ref{alg:pseudo-label-gmm} and~\ref{alg:t_weight}.}

\begin{minipage}[t]{0.48\textwidth}
\begin{algorithm}[H]
  \caption{Pseudo-label target GMM.}
  \label{alg:pseudo-label-gmm}
  \Function{\textcolor{black}{propagate\_labels($\mathbf{X}^{(P_{S})}, \mathbf{Y}^{(P_{S})}, \mathbf{X}^{(P_{T})}$)}}{
    $\textcolor{black}{P \leftarrow \text{CondtionalEM}(\mathbf{X}^{(P_{S})}, \mathbf{Y}^{(P_{S})})}$\;
    $\textcolor{black}{Q \leftarrow \text{EM}(\mathbf{X}^{(P_{T})})}$\;
    $\textcolor{black}{\omega \leftarrow \text{GMMOT}(P_{S}, P_{T})}$\;
    $\textcolor{black}{\nu^{(P_{T})} = \nicefrac{\omega^{T}\nu^{(P_{S})}}{\pi^{(P_{T})}}}$\;
    \Return{\textcolor{black}{$\nu^{(P_{T})}$}}\;
  }
\end{algorithm}
\end{minipage}
\hfill
\begin{minipage}[t]{0.48\textwidth}
\begin{algorithm}[H]
  \caption{$T_{weight}$.}
  \label{alg:t_weight}
  \Function{\textcolor{black}{$T_{weight}$($\mathbf{x}^{(P_{S})}, y^{(P_{S})}, P_{S}, \omega, \tau$)}}{
    \Comment{Using Equation~\ref{eq:component_estimation}}
    $\textcolor{black}{k_{1} \leftarrow \text{estimate\_components}(\mathbf{x}^{(P_{S})}, P_{S})}$\;

    \For{\textcolor{black}{$j$ such that $\omega_{k_{1}, j} \geq \tau$}}{
        \Comment{Using eq.~\ref{eq:gauss_map} or ~\ref{eq:axis_aligned_gauss}}
        $\textcolor{black}{\tilde{\mathbf{x}}_{k_{2}}^{(P_{S})} \leftarrow T_{k_{1}, k_{2}}(\mathbf{x}^{(P_{S})})}$\;

        $\textcolor{black}{\tilde{y}_{k_{2}}^{(P_{S})} \leftarrow y^{(P_{S})}}$

        $\textcolor{black}{w_{k_{2}} \leftarrow \omega_{k_{1}, k_{2}}}$
    }
    
    \Return{$\textcolor{black}{\{w_{k_{2}}, \tilde{\mathbf{x}}_{k_{2}}^{(P)}, y_{k_{2}}^{(P)}\}_{k_{2}:\omega_{k_{1},k_{2}} \geq \tau}}$}\;
  }
\end{algorithm}
\end{minipage}

\noindent\textcolor{black}{\noindent\textbf{Limitations.} In this work, we assume that data is multi-modal, which is often the case in classification. Furthermore, we assume that it can be modeled accurately through \glspl{gmm}, i.e., data is well separated into clusters. While we can expect estimation and inference to be difficult in high-dimensions, we show in our experiments that our method outperform previous baselines based on empirical \gls{ot}. Finally, we show in our appendix that our methods are robust to the \emph{over estimation} of \gls{gmm} components.}

\subsection{Label Propagation and Maximum a Posterior Estimation}

\begin{figure}[t]
    \centering
    \begin{subfigure}[t]{0.23\linewidth}
        \includegraphics[width=\linewidth]{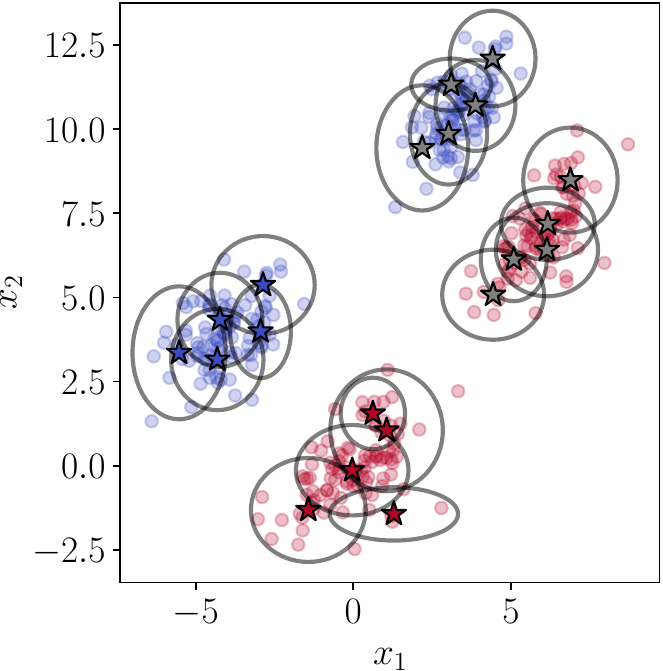}
        \caption{}
    \end{subfigure}\hfill
    \begin{subfigure}[t]{0.23\linewidth}
        \includegraphics[width=\linewidth]{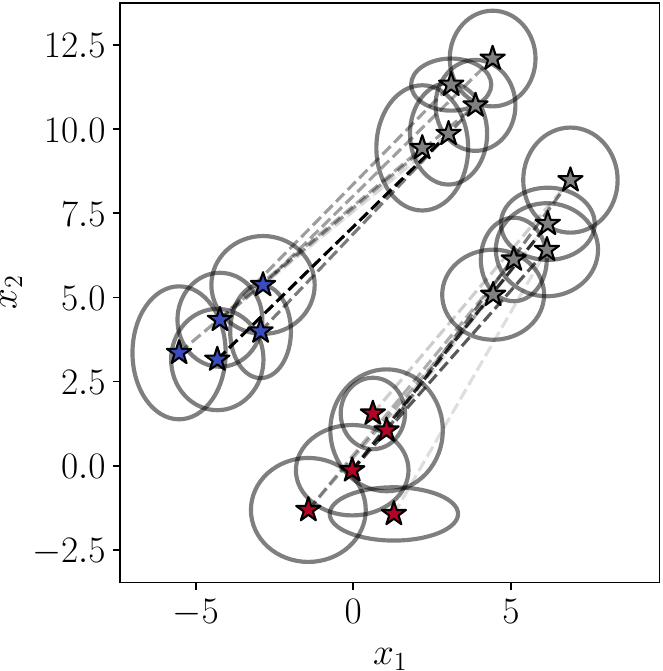}
        \caption{}
    \end{subfigure}\hfill
    \begin{subfigure}[t]{0.23\linewidth}
        \includegraphics[width=\linewidth]{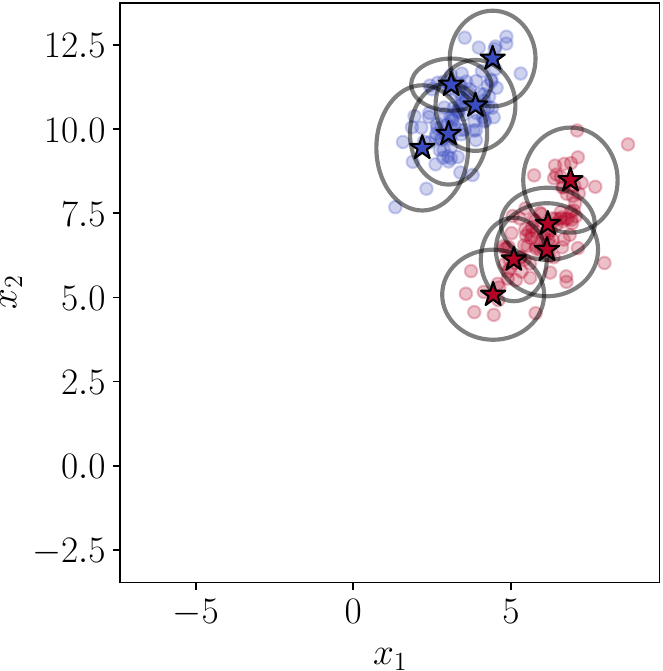}
        \caption{}
    \end{subfigure}\hfill
    \begin{subfigure}[t]{0.23\linewidth}
        \includegraphics[width=\linewidth]{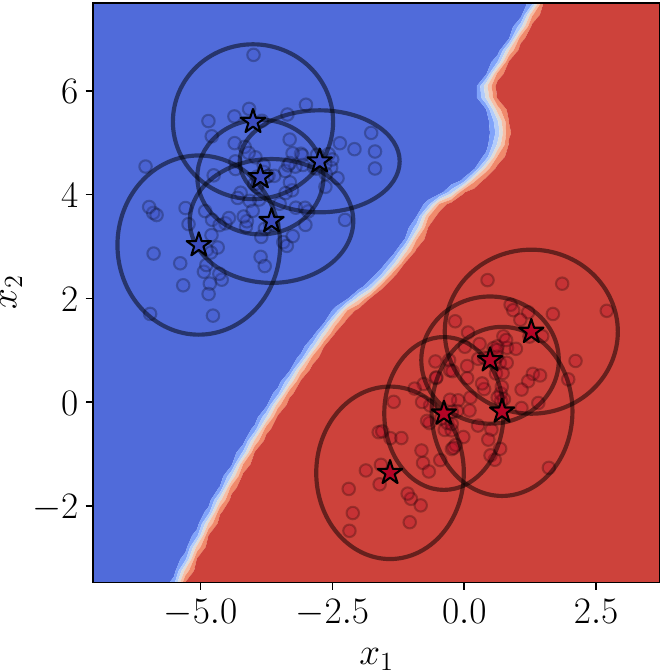}
        \caption{}
    \end{subfigure}
    \caption{\textbf{Illustration of the label propagation strategy.} (a) Shows the source and target \glspl{gmm}. \glspl{gmm} are represented through their means (stars) and covariance matrices (ellipses). The labels of components are represented through colors. Since the target domain is unlabeled, their means are gray colored. (b) Shows the \gls{ot} plan between components, $\omega$. $\omega$ allows us to propagate the labels of source domain \gls{gmm} towards the target. (c) Shows the obtained target \gls{gmm}. Finally, (d) shows the \gls{map} classifier.}
    \label{fig:label_prop}
\end{figure}

Recall that, in equation~\ref{eq:GMMOT}, the result of the \gls{gmmot} problem is a transportation plan $\omega$, between components of the \glspl{gmm} $P$ and $Q$. \textcolor{black}{As a result, $\omega$ has as marginals the probability vectors $\pi^{(P)}$ and $\pi^{(Q)}$. Furthermore, in the original probabilistic view of \glspl{gmm}, $\pi_{k_{1}}^{(P_{S})} = \pr(K_{S}=k_{1})$, whereas $\pi_{k_{2}}^{(P_{T})} = \pr(K_{T}=k_{2})$. Given this interpretation, it is natural to see the transportation plan as $\omega_{k_{1},k_{2}} = \pr(K_{S}=k_{1},K_{T}=k_{2})$. Note that we can estimate the probability $\pr(Y|K_{T})$ using the law of total probabilities,}
\begin{align*}
    \textcolor{black}{\pr(Y|K_{T}=k_{2}) = \sum_{k_{1}=1}^{K_{S}}\pr(Y|K_{S}=k_{1},K_{T}=k_{2})\pr(K_{S}=k_{1}|K_{T}=k_{2}).}
\end{align*}
Here, assuming that $Y$ and $K_{2}$ are conditionally independent given $K_{1}$, we have,
\begin{align*}
    \textcolor{black}{\pr(Y|K_{T}=k_{2}) = \sum_{k_{1}=1}^{K_{S}}\pr(Y|K_{S}=k_{1})\pr(K_{S}=k_{1}|K_{T}=k_{2}),}
\end{align*}
This assumption plays the same role as covariate shift hypothesis~\cite{sugiyama2007covariate} in conventional \gls{da} works. On an intuitive level, our assumption explicit the fact that $K_{T}$ is redundant with respect to $K_{S}$. The conditional \textcolor{black}{$\pr(K_{S}|K_{T}) = \nicefrac{\pr(K_{S},K_{T})}{\pr(K_{T})}$} can be computed through the optimal transport plan, i.e.,
\begin{align}
    \textcolor{black}{\hat{\nu}_{k_{2}}^{(P_{T})} = \dfrac{1}{\pi_{k_{2}}^{(P_{T})}}\sum_{k_{1}=1}^{K_{S}}\omega_{k_{1},k_{2}}\nu_{k_{1}}^{(P_{S})}\text{, or, }\hat{\nu}^{(P_{T})} = \dfrac{\omega^{T}\nu^{(P_{S})}}{\pi^{(Q_{T})}},}\label{eq:label_propagation}
\end{align}
where the division should be understood elementwise. Equation~\ref{eq:label_propagation} is known in the \gls{ot} literature as \emph{label propagation}~\cite{redko2019optimal}, and, as we discussed in the related works section, has been used extensively in the context of empirical \gls{ot}. Given the estimated labels $\textcolor{black}{\hat{\nu}_{k_{2}}^{(P_{T})}}$, we effectively defined a labeled \gls{gmm} for the target domain. Based on this \gls{gmm}, we can perform \gls{map} estimation to define a classifier in the target domain,
\begin{align}
    \textcolor{black}{\hat{h}_{MAP}(\mathbf{x})} &= \argmax{y=1,\cdots,n_{c}}\pr(Y=y|X=\mathbf{x}),\nonumber\\
    &= \argmax{y=1,\cdots,n_{c}}\sum_{k=1}^{K_{T}}\pr(K_{T}=k|X=\mathbf{x})\pr(Y=y|X=\mathbf{x},K_{T}=k),\nonumber\\
    &= \textcolor{black}{\argmax{y=1,\cdots,n_{c}}\sum_{k=1}^{K_{T}}\biggr(\dfrac{\pi_{k}^{(P_{T})}P_{T,k}(\mathbf{x}) }{ \sum_{k'=1}^{K_{T}}\pi_{k'}^{(P_{T})}P_{T,k'}(\mathbf{x}) }\biggr)\nu_{k}^{(P_{T})}},\label{eq:gmm_otda_map}
\end{align}
where, from the second to the third equality we assumed that $Y$ and $K$ are conditionally independent given $X$. This hypothesis is intuitive, as classes and components are representing the same structures within the data points. We show an illustration of these ideas in Figure~\ref{fig:label_prop}.

\subsection{Mapping Estimation}

In this section, we propose a new mapping estimation technique between two \glspl{gmm}, \textcolor{black}{$P_{S}$ and $P_{T}$}. As discussed in~\cite[Section 6.3]{delon2020wasserstein},  this problem is not straightforward. Indeed, since $\gamma$ is a \gls{gmm}, it cannot be, in general, written as \textcolor{black}{$(Id, T)_{\sharp}P_{S}$}. Thus, these authors consider two strategies: a mean map and a random map. First, samples are mapped \textcolor{black}{$T_{mean}(\mathbf{x}^{(P_{S})}) = \mathbb{E}_{\mathbf{x}^{(P_{T})} \sim \gamma(\cdot|\mathbf{x}^{(P_{S})})}[\mathbf{x}^{(P_{T})}]$}, but it may end up not actually matching \textcolor{black}{$P_{S}$} with \textcolor{black}{$P_{T}$} (see, e.g.,~\cite[Section 6.3]{delon2020wasserstein}).

Second,~\cite{delon2020wasserstein} defines
\begin{align*}
\textcolor{black}{T_{rand}(\mathbf{x}^{(P_{S})}) = T_{k_{1},k_{2}}(\mathbf{x}^{(P_{S})})\text{ with probability }p_{k_{1},k_{2}}(\mathbf{x}^{(P_{S})}) = \omega_{k_{1},k_{2}}^{\star}\dfrac{\mathcal{N}(\mathbf{x}^{(P_{S})}|\mu_{k_{1}}^{(P_{S})}, \Sigma_{k_{1}}^{(P_{S})})}{\sum_{k}p_{k}\mathcal{N}(\mathbf{x}^{(P_{S})}|\mu_{k}^{(P_{S})}, \Sigma_{k}^{(P_{S})})},}
\end{align*}
which has the advantage of matching \textcolor{black}{$P_{S}$} with \textcolor{black}{$P_{T}$}. However, as noted by~\cite{delon2020wasserstein}, $T_{rand}$ usually leads to irregular mappings due the sampling procedure of indices $(k_{1}, k_{2})$ as shown in~\cite[Figure 7]{delon2020wasserstein}.

We put forth a third strategy for mapping $P_{S}$ into $P_{T}$. Our intuition is twofold. First, we can increase the regularity of $T_{rand}$, by estimating the component $k_{1}$ that most likely originated $\mathbf{x}^{(P_{S})}$, that is,
\begin{align}
    \textcolor{black}{k_{1} := \argmax{k=1,\cdots,K_{S}}\pr(K_{S}=k|X_{S}=\mathbf{x}^{(P_{S})}) = \dfrac{\pi^{(P_{S})}_{k}P_{S,k}(\mathbf{x}^{(P_{S})})}{\sum_{k'=1}^{K_{S}}\pi^{(P_{S})}_{k'}P_{S,k'}(\mathbf{x}^{(P_{S})})}.}\label{eq:component_estimation}
\end{align}
Second, we map \textcolor{black}{$\mathbf{x}^{(P_{S})}$} into the components of \textcolor{black}{$P_{T}$}. Note that, since the marginals \textcolor{black}{$\pi^{(P_{S})}$} and \textcolor{black}{$\pi^{(P_{T})}$} are different, the optimal transport plan $\omega$ may split the mass of \textcolor{black}{$P_{S,k_{1}}$} into several \textcolor{black}{$P_{T,k_{2}}$}. As a result, we produce \textcolor{black}{$\{T_{k_{1},k_{2}}(\mathbf{x}^{(P_{S})})\}_{k_{2}:\omega_{k_{1},k_{2}} \geq \tau}$}, i.e., we map $\mathbf{x}^{(P_{S})}$ to all components \textcolor{black}{$P_{T,k_{2}}$} such that $\omega_{k_{1},k_{2}} \geq \tau \geq 0$. In principle, one may choose $\tau = 0$ and filter only the components that are not matched with \textcolor{black}{$P_{S,k_{1}}$}.  Third, we further weight the importance of generated samples, by using $\omega_{k_{1},k_{2}}$ as sample weights. At the end, we generate a weighted dataset \textcolor{black}{$\{ (\omega_{k_{1}, k_{2}}, T_{k_{1},k_{2}}(\mathbf{x}_{i}^{(P_{S})}), y_{i}^{(P_{S})}) \}_{i=1}^{m}$}, where $m$ is the total amount of samples generated. We call our overall mapping $T_{weight}$, for which a pseudo-code is presented in~\ref{alg:t_weight}.

The mapping we just defined has a few interesting properties. First, it is a piece-wise affine map, as each $T_{k_{1},k_{2}}$ is affine. This property contrast with the Gaussian hypothesis, which defines an affine map between \textcolor{black}{$P_{S}$} and \textcolor{black}{$P_{T}$}. Second, with respect the transportation of samples, our mapping strategy is naturally \emph{group-sparse}, in the sense of~\cite{courty2016otda}. This claim comes from the fact that samples in $P_{S}$ are transported based on the Gaussian component they belong to. Third, our mapping is defined on the whole support of the \gls{gmm} \textcolor{black}{$P_{S}$}. In contrast, empirical \gls{ot} is only defined on the samples \textcolor{black}{$\{\mathbf{x}_{i}^{(P_{S})}\}_{i=1}^{n}$} of \textcolor{black}{$P_{S}$}.
\begin{figure}[t]
    \centering
    \begin{subfigure}[t]{0.3\linewidth}
        \includegraphics[width=\linewidth]{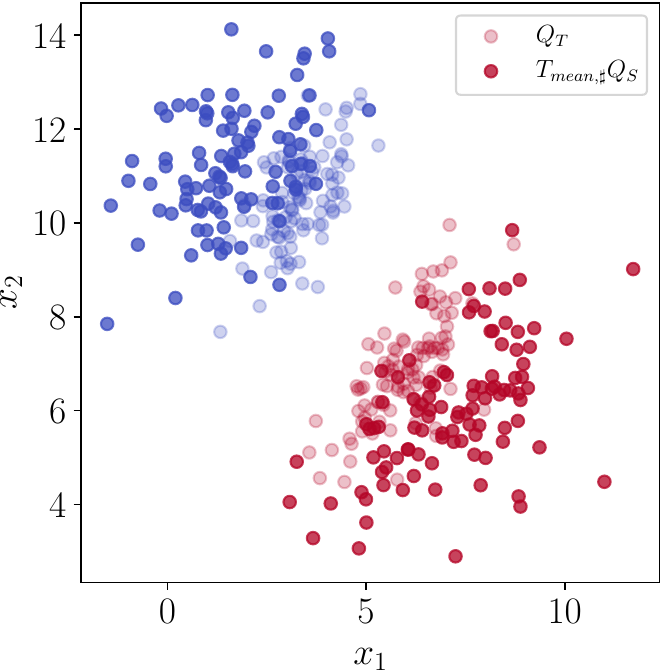}
        \caption{$T_{mean}$.}
    \end{subfigure}\hfill
    \begin{subfigure}[t]{0.3\linewidth}
        \includegraphics[width=\linewidth]{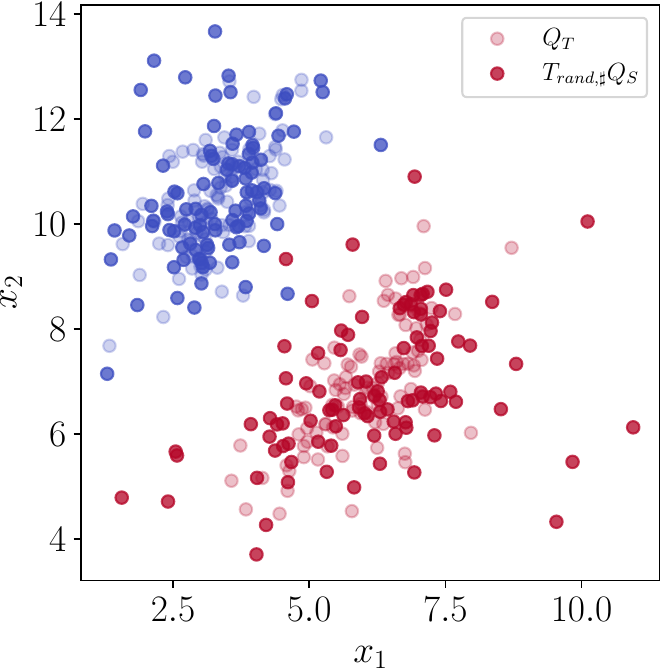}
        \caption{$T_{rand}$.}
    \end{subfigure}\hfill
    \begin{subfigure}[t]{0.3\linewidth}
        \includegraphics[width=\linewidth]{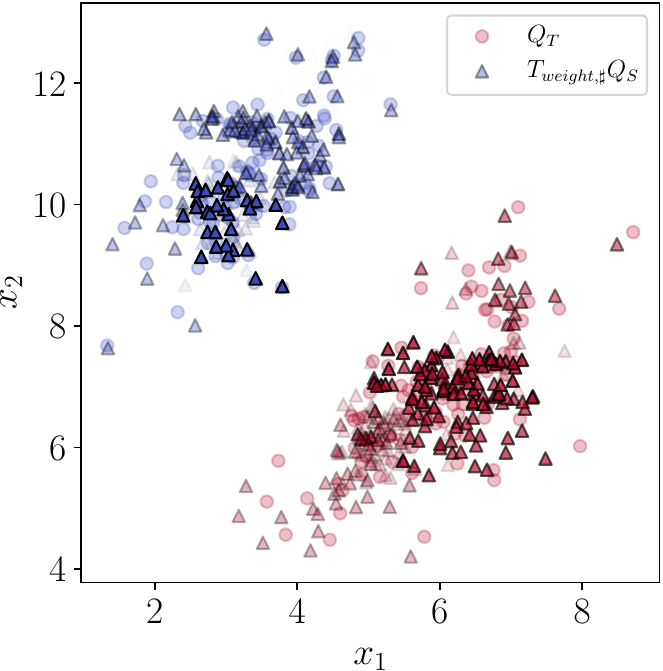}
        \caption{$T_{weight}$.}
    \end{subfigure}
    \caption{\textbf{Mapping estimation using \gls{gmmot}.} In (a) and (b), we show the $T_{mean}$ and $T_{rand}$ strategies of~\cite{delon2020wasserstein}, respectively. In (c), we show our strategy $T_{weight}$. Our mapping reduces randomness by first estimating the component $k_{1}$ most likely to have generated $\mathbf{x}^{(P)}$. Then, we weight the importance of transported samples by $\omega_{k_{1},k_{2}}$.}
    \label{fig:enter-label}
\end{figure}

\section{Experiments}\label{sec:experiments}

\textcolor{black}{In this section, we present our experiments with \gls{da}. We consider a wide range of 9 benchmarks in computer vision and fault diagnosis. In the first case, we consider Caltech-Office~\cite{gong2012geodesic}, ImageCLEF~\cite{caputo2014imageclef}, Office31~\cite{saenko2010adapting}, Office-Home~\cite{venkateswara2017deep}, (MNIST, USPS, SVHN)~\cite{seguy2017large} and VisDA~\cite{peng2017visda}. In the second case, we consider the CWRU\footnote{\url{https://engineering.case.edu/bearingdatacenter}}, CSTR~\cite{pilario2017process,montesuma2022cross} and TE process benchmarks~\cite{montesuma2024benchmarking}. Further details on these benchmarks are available in the appendix. In our experiments, an adaptation task is a pair $(S, T)$ of a source domain $S$ and a target domain $T$. To summarize our experimentation, there are in total 9 benchmarks, and 85 domain adaptation tasks.}

\textcolor{black}{For computer vision benchmarks, we follow previous research~\cite{el2022hierarchical,chuang2023infoot} and pre-train ResNet networks~\cite{he2016deep} on the source domains. We then use the encoder branch as a feature extractor, and perform \emph{shallow \gls{da}} on the extracted features. These feature serve as the basis for each domain adaptation algorithm. With the exception of GMM-OTDA$_{MAP}$, performance on the target domain is based on the generalization of a 1-layer neural network trained with transformed data. For GMM-OTDA$_{MAP}$ we use the \gls{map} strategy described in equation~(\ref{eq:gmm_otda_map}). For the \emph{MNIST, USPS, SVHN} benchmark, we follow~\cite{seguy2017large} to obtain comparable results. For $M \rightarrow U$ and $U\rightarrow M$ we downsize MNIST to the resolution of USPS, i.e., we downscale images to $(16, 16)$. For $M \rightarrow S$, we upscale MNIST images to match the resolution of SVHN, i.e., $(32, 32)$, then we use features extracted from the last layer of a LeNet5. We refer readers to~\cite{seguy2017large} and~\cite{struckmeier2023learning} for further details on these benchmarks. }

\textcolor{black}{Our main point of comparison is with other \gls{ot}-based \gls{da} methods. We compare our GMM-OTDA strategies with other \gls{ot} methods for \gls{da}, namely, we consider the \gls{otda} strategy of~\cite{courty2016otda} (Exact and Sinkhorn), the linear mapping estimation of~\cite{flamary2019concentration}, and the Info\gls{ot} strategy of~\cite{chuang2023infoot} (barycentric and conditional mappings). For the scalability experiment using the \emph{MNIST, USPS, SVHN} benchmark, we consider the large scale \gls{ot} methods of~\cite{seguy2017large}, denoted as Alg. 1 and 2. Furthermore, for completeness, we consider the \gls{laot} strategy of~\cite{struckmeier2023learning}.}

In the following, we divide our experiments into 6 sections. Section~\ref{sec:better} figures experiments with high-dimensional \gls{da} problems. Section~\ref{sec:scalability} shows experiments with large-scale \gls{da} datasets. Section~\ref{sec:visda} shows our results on the VisDA-C benchmark. Section~\ref{sec:cdfd} focuses on our experiments with \gls{cdfd}. Section~\ref{sec:ablations} performs an ablation study, as well as the visualization of adaptation in the \gls{cwru} benchmark. Finally, section~\ref{sec:runtime} shows a running time comparison with the tested algorithms.

\subsection{Scalability with respect $d$}\label{sec:better}

In this section, our goal is to evaluate how our method scales with the data dimensionality $d$. To do so, we evaluate methods based on their performance on visual adaptation benchmarks. The goal is to classify images into categories, based on $2048-$dimensional vectors from ResNets~\cite{he2016deep} fine-tuned on the source domain of each adaptation task. We summarize our results in Figure~\ref{fig:ssda_aggregated}. The detailed results may be found in the appendix, i.e., Table~\ref{tab:single_source_results}.

Over Caltech-Office and ImageCLEF, our methods outperform other state-of-the-art methods. For Office 31 and Office-Home, Info-OT$_{c}$ and OTDA$_{affine}$ proved to be more effect than our methods, but ours still ranks second. For Office 31, the density estimation strategy of Info-OT$_{c}$ proves effective in finding a better map between the domains. For Office-Home, the baseline is already one of the best performing methods. An affine transformation is therefore sufficient for an effective adaptation.

Nonetheless, our method surpasses empirical \gls{ot} over all tested benchmarks, especially preventing negative transfer in the Office-Home benchmark. Likewise, approximating class conditional distributions $P(X|Y)$ through \glspl{gmm} proves effective over empirical \gls{ot}. Indeed, our method improves over HOT-DA of~\cite{el2022hierarchical}, which is based on empirical distributions.

These experiments prove that our method can effectively perform \gls{uda} between high-dimensional distributions. Note that, for these benchmarks, we use $2048-$dimensional vectors, which is by far the largest dimensionality values considered in this study.

\begin{figure}[ht]
    \centering
    \includegraphics[width=\linewidth]{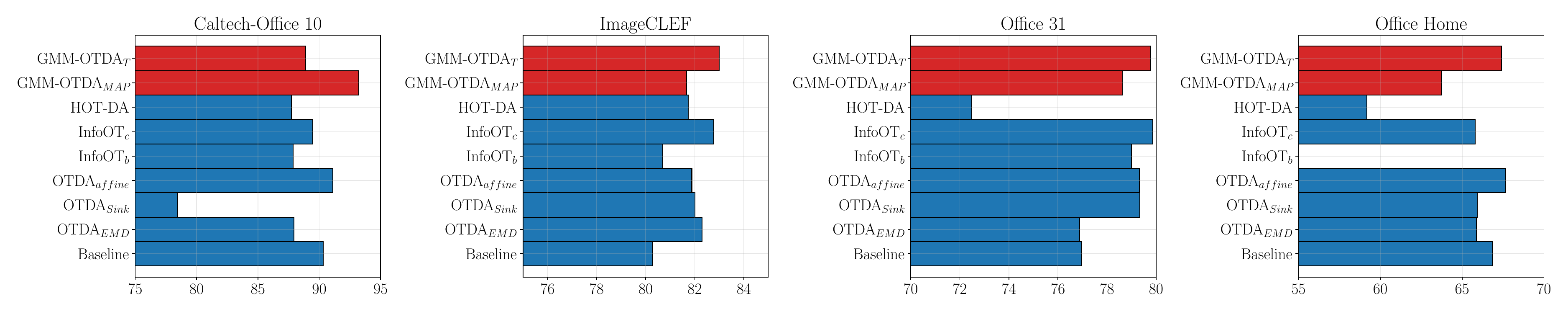}
    \caption{Average adaptation performance over $4$ visual domain adaptation benchmarks.}
    \label{fig:ssda_aggregated}
\end{figure}

\subsection{Scalability with resepct $n$}\label{sec:scalability}

In this section, we use the \emph{MNIST, USPS, SVHN} benchmark. Our goal is to evaluate how our method scales with the number of samples $n$. As we discussed in Remark~\ref{remark:complexity}, a major advantage of the \gls{gmm} formulation is reducing \gls{ot} complexity from $\mathcal{O}(n^{3}\log n)$ to $\mathcal{O}(K^{3}\log K)$, i.e., we replace number of samples by number of components, which are orders of magnitude inferior. To give a practical comparison, while MNIST has $n = 6\times 10^{5}$ samples, we represent its probability distribution through a \gls{gmm} with $K = 10^{2}$ components. This modeling choice improves scalability, especially when the components of \glspl{gmm} have diagonal covariance matrices.

With respect performance, table~\ref{tab:mnist_svhn_usps} shows that GMM-OTDA has a similar performance to \gls{laot}, i.e., performance degrades on $M \rightarrow U$, but increases on $U \rightarrow M$ and $M \rightarrow S$. Curiously, this corresponds to the case where a simpler dataset (e.g., MNIST) is transferred to a more complex dataset (e.g., SVHN). The performance similarity is not surprising, since \gls{laot} and GMM-OTDA work under similar principles. However, the \gls{gmm} modeling, again, proves superior to the Gaussian hypothesis, as we are able to improve performance on $U \rightarrow M$ and $M \rightarrow S$ tasks.

\begin{table}[ht]
    \centering
    \begin{tabular}{lccccccccc}
        \toprule
        Algorithm & $M\rightarrow U$ & $U\rightarrow M$ & $M \rightarrow S$\\
        \midrule
        Baseline & 73.47 & 36.97 & 54.33\\
        OTDA$_{EMD}$ & 57.75 & 52.46 & -\\
        OTDA$_{Sink}$ & 68.75 & 57.35 & -\\
        \textcolor{black}{Alg. 1 of~\cite{seguy2017large} with $H$ }& 68.84 & 57.55 & 58.87 \\
        \textcolor{black}{Alg. 1 of~\cite{seguy2017large} with $\ell_{2}$.} & 67.80 & 57.47 & 60.56\\
        \textcolor{black}{Alg. 1 + 2 of~\cite{seguy2017large} with $H$ }& \textbf{77.92} & 60.02 & 61.11\\
        \textcolor{black}{Alg. 1 + 2 of~\cite{seguy2017large} with $\ell_{2}$.} & 72.61 & 60.50 & 62.88 \\
        LaOT & 72.57 & 62.28 & 60.36\\
        \midrule
        GMM-OTDA (ours) & 71.83 & \textbf{63.11} & \textbf{87.19}\\
        \bottomrule
    \end{tabular}
    \caption{Large scale \gls{ot} experiment. We consider the adaptation between 3 digit recognition benchmarks, namely, USPS, MNIST and SVHN. Overall, GMM-OTDA largely outperforms other methods on harder adaptation tasks, i.e., $U \rightarrow M$ and $M \rightarrow S$.}
    \label{tab:mnist_svhn_usps}
\end{table}

\subsection{VisDA-C Benchmark}\label{sec:visda}

\textcolor{black}{We experiment with the VisDA benchmark~\cite{peng2017visda}, a large scale \gls{da} dataset containing 152397 and 55388 source and target domain samples. As in the previous benchmarks, we pre-train the feature extractor using source domain data, then proceed to perform adaptation over the extracted features. This experiment stresses the scalability of our strategy in comparison with empirical \gls{ot} methods, since solving an \gls{ot} problem over this benchmark would lead to a linear program with $n_{S} \times n_{T} = 8.44 \times 10^{9}$ variables. More dramatically, for running the barycentric map over this benchmark one would need to store $\gamma$, leading to $n_{S}\times n_{T}$ floating point coefficients, that is, approximately $270.11$ GB of memory.}

\textcolor{black}{To cope with the sheer volume of data, we run empirical \gls{ot} methods on a sub-sample of $n_{S} = n_{T} = 15000$ samples. Parametric versions of \gls{ot} methods, such as OTDA$_{affine}$ and GMM-OTDA are run with the full datasets, which illustrates the advantage of having a compact representation for distributions. Besides, we explore how these methods improve performance over different feature extractors. Hence, besides ResNet 50 and 101, we also consider a ViT-16-b~\cite{dosovitskiy2021an}. Our results are shown in table~\ref{tab:algorithm_architectures}.}

\begin{table}[ht]
\centering
\begin{tabular}{lccc}
\toprule
\textbf{Algorithm} & \textbf{ResNet 50} & \textbf{ResNet101} & \textbf{ViT-b-16} \\ 
\midrule
Source-Only & 47.93 & 53.90 & 56.70 \\ 
OTDA$_{\text{EMD}}$ & 53.69 {\scriptsize \color{pastelgreen}($\Delta +5.76$)} & 57.42 {\scriptsize \color{pastelgreen}($\Delta +3.52$)} & 63.25 {\scriptsize \color{pastelgreen}($\Delta +6.55$)} \\ 
OTDA$_{\text{Sinkhorn}}$ & 53.02 {\scriptsize \color{pastelgreen}($\Delta +5.09$)} & 10.54 {\scriptsize \color{pastelred}($\Delta -43.36$)} & 66.75 {\scriptsize \color{pastelgreen}($\Delta +10.05$)} \\ 
OTDA$_{\text{Affine}}$ & 7.46 {\scriptsize \color{pastelred}($\Delta -40.47$)} & 11.82 {\scriptsize \color{pastelred}($\Delta -42.08$)} & 6.75 {\scriptsize \color{pastelred}($\Delta -49.95$)} \\ 
OTDA$_{\text{Affine-Diag}}$ & 51.41 {\scriptsize \color{pastelgreen}($\Delta +3.48$)} & 56.91 {\scriptsize \color{pastelgreen}($\Delta +3.01$)} & 59.94 {\scriptsize \color{pastelgreen}($\Delta +3.24$)} \\ 
HOTDA & 47.51 {\scriptsize \color{pastelred}($\Delta -0.42$)} & 47.64 {\scriptsize \color{pastelred}($\Delta -6.26$)} & 62.55 {\scriptsize \color{pastelgreen}($\Delta +5.85$)} \\ 
GMM-OTDA$_{\text{T}}$ & \textbf{58.35} {\scriptsize \color{pastelgreen}($\Delta +10.42$)} & 56.57 {\scriptsize \color{pastelgreen}($\Delta +2.67$)} & 74.10 {\scriptsize \color{pastelgreen}($\Delta +17.40$)} \\ 
GMM-OTDA$_{\text{MAP}}$ & 57.36 {\scriptsize \color{pastelgreen}($\Delta +9.43$)} & \textbf{58.77} {\scriptsize \color{pastelgreen}($\Delta +4.87$)} & \textbf{74.44} {\scriptsize \color{pastelgreen}($\Delta +17.74$)} \\ 
\bottomrule
\end{tabular}
\caption{\textcolor{black}{Comparison of domain adaptation performance over different feature extractors pre-trained with source domain data. We report the classification accuracy (in \%) and the difference $\Delta$ over the source-only baseline. Our methods GMM-OTDA T and MAP consistently outperform other OT-based methods.}}
\label{tab:algorithm_architectures}
\end{table}

\textcolor{black}{From table~\ref{tab:algorithm_architectures}, note that our GMM-OTDA methods consistently outperform other OT-based methods. Furthermore, we ablate on the choice of using diagonal covariances, by comparing the performance of estimating an \gls{ot} map between Gaussian measures with full (equation~\ref{eq:gauss_map}) and diagonal covariances (equation~\ref{eq:axis_aligned_gauss}). Hence, using diagonal covariances provide a regularizing effect, improving the estimation of an \gls{ot} map between Gaussian measures when the full-covariances are singular.}

\begin{figure}[t]
    \centering
    \begin{subfigure}{0.32\linewidth}
        \includegraphics[width=\linewidth]{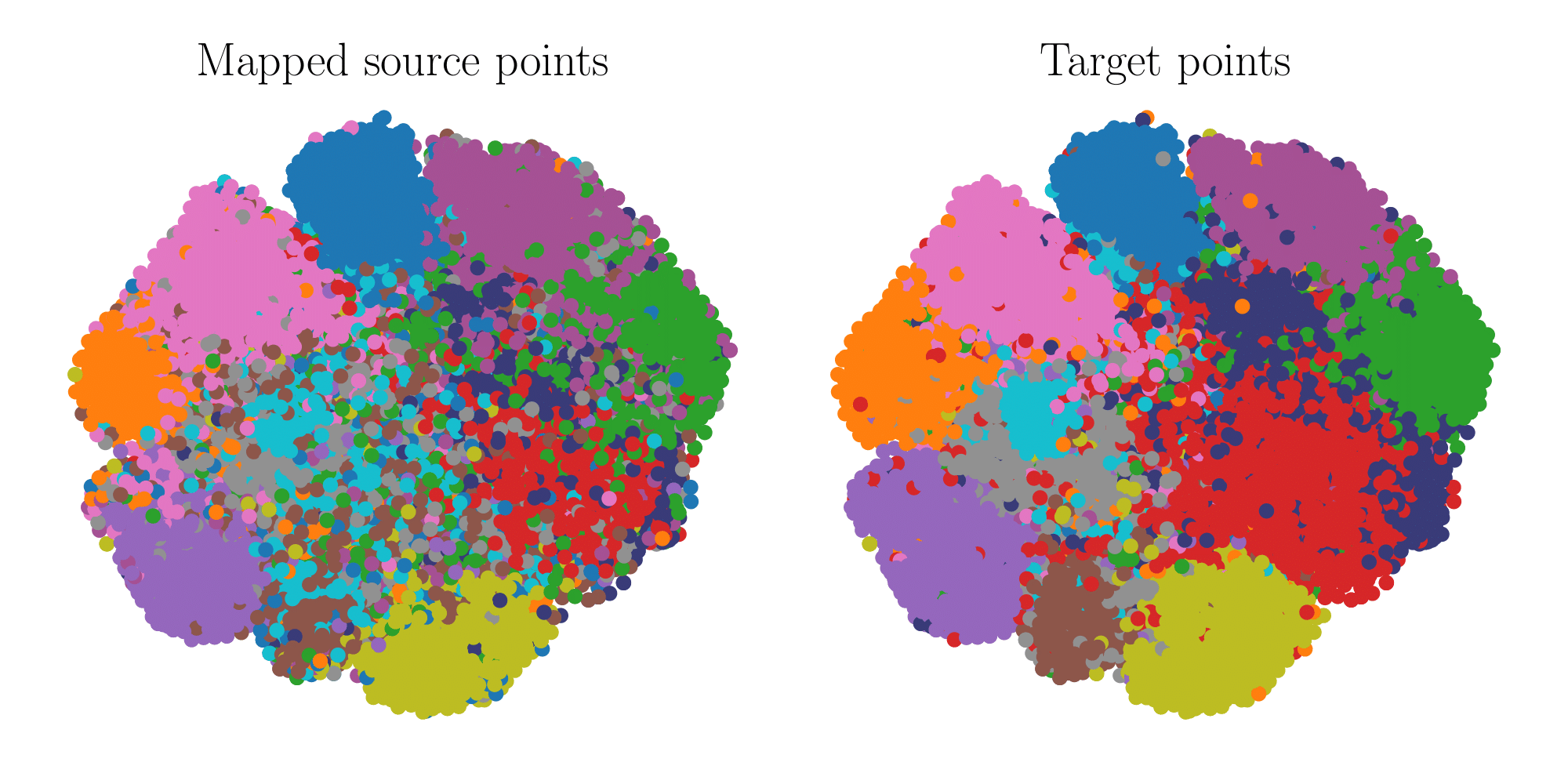}
        \caption{OTDA$_{EMD}$}
    \end{subfigure}\hfill
    \begin{subfigure}{0.32\linewidth}
        \includegraphics[width=\linewidth]{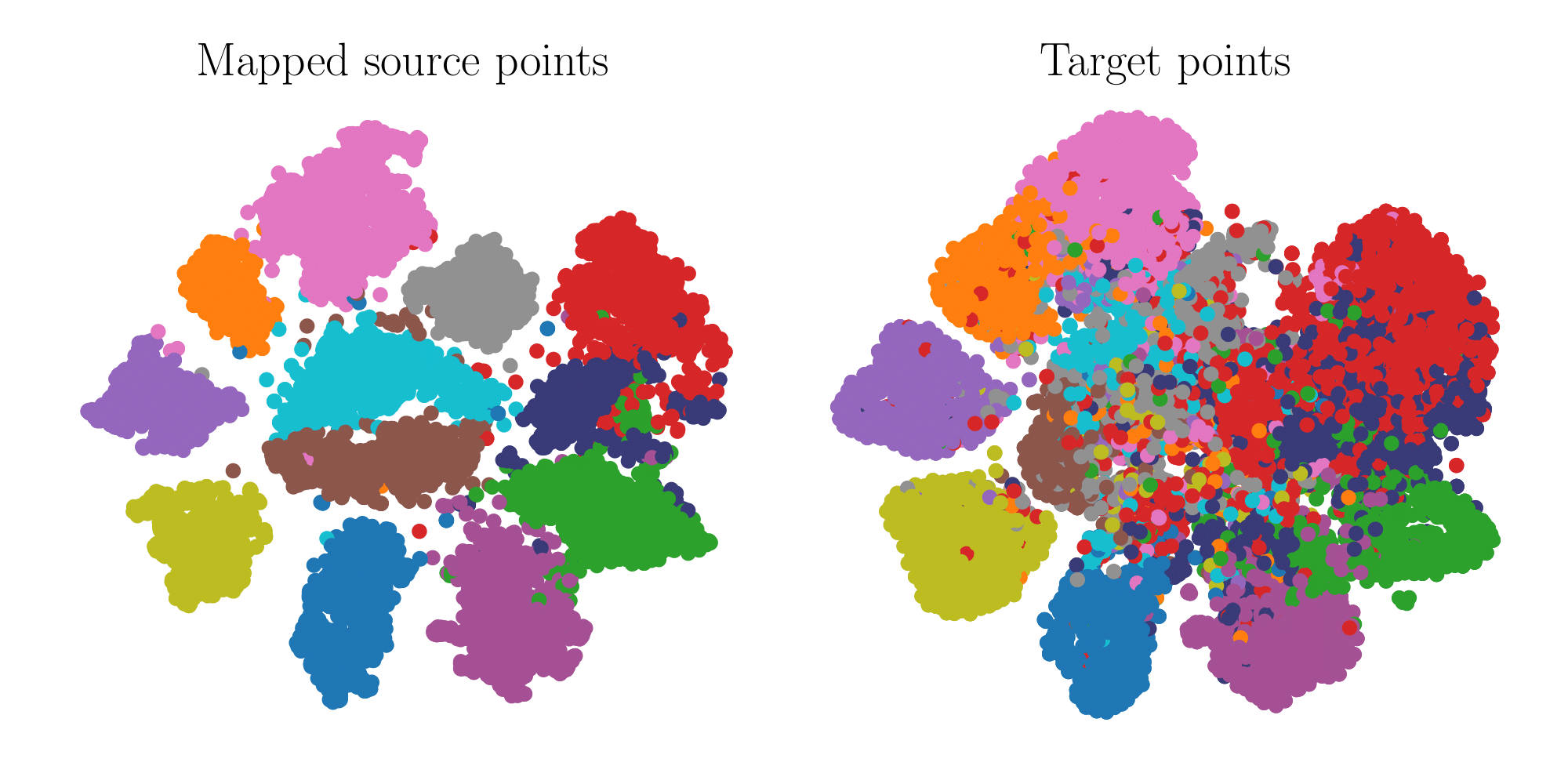}
        \caption{OTDA$_{Sink}$}
    \end{subfigure}\hfill
    \begin{subfigure}{0.32\linewidth}
        \includegraphics[width=\linewidth]{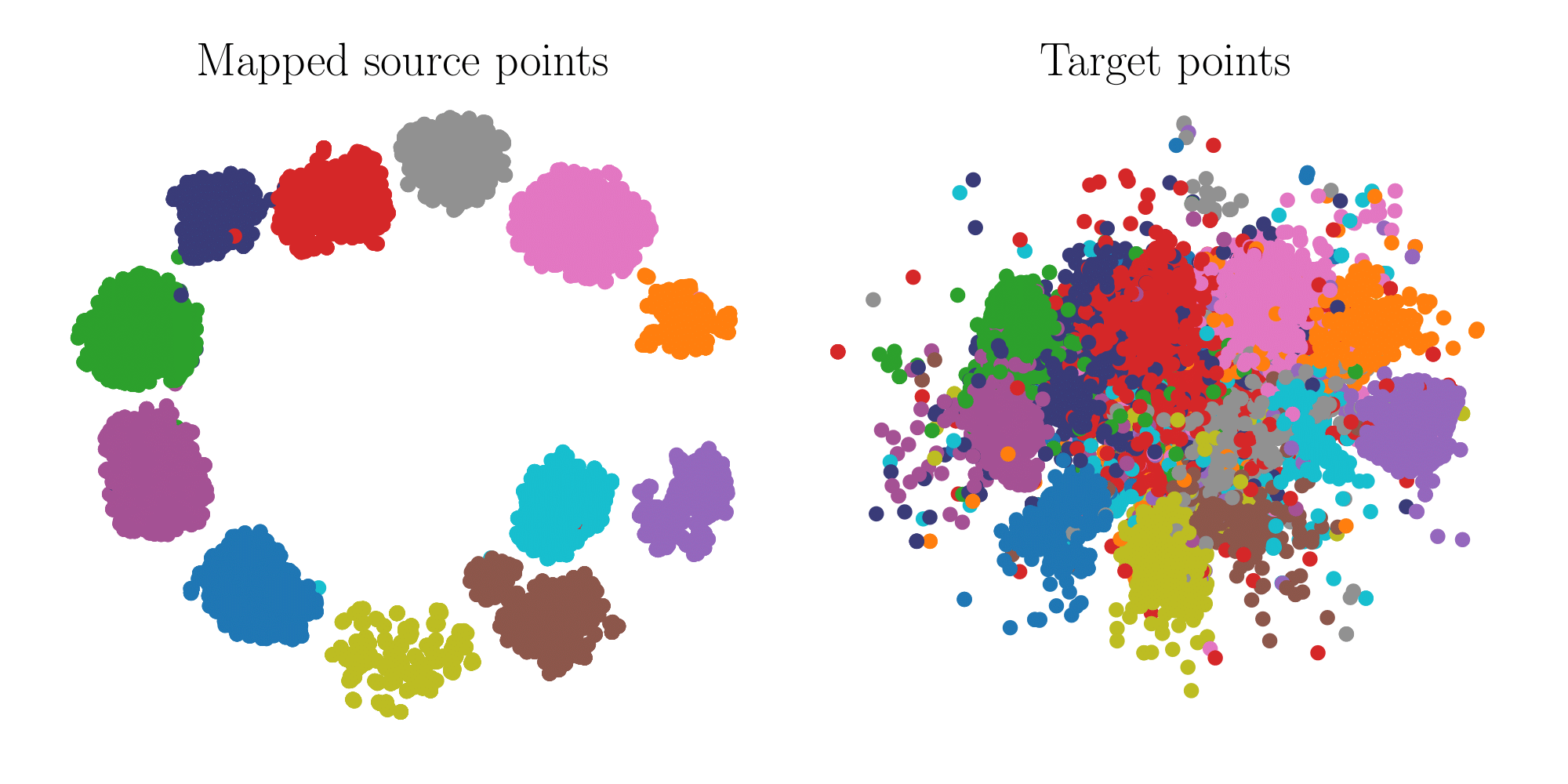}
        \caption{OTDA$_{affine}$}
    \end{subfigure}\hfill\\
        \begin{subfigure}{0.32\linewidth}
        \includegraphics[width=\linewidth]{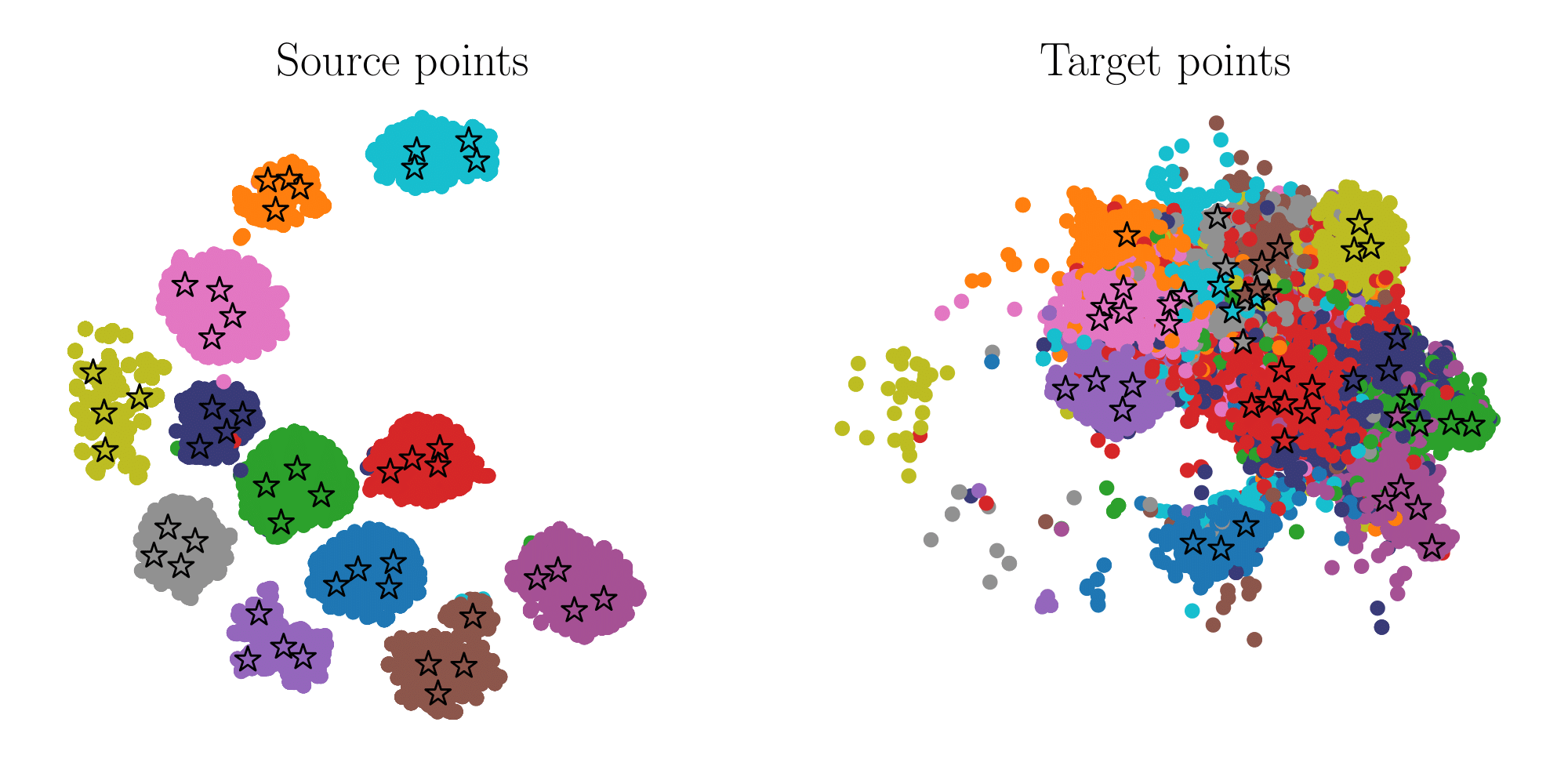}
        \caption{GMM-OTDA$_{MAP}$}
    \end{subfigure}
    \begin{subfigure}{0.32\linewidth}
        \includegraphics[width=\linewidth]{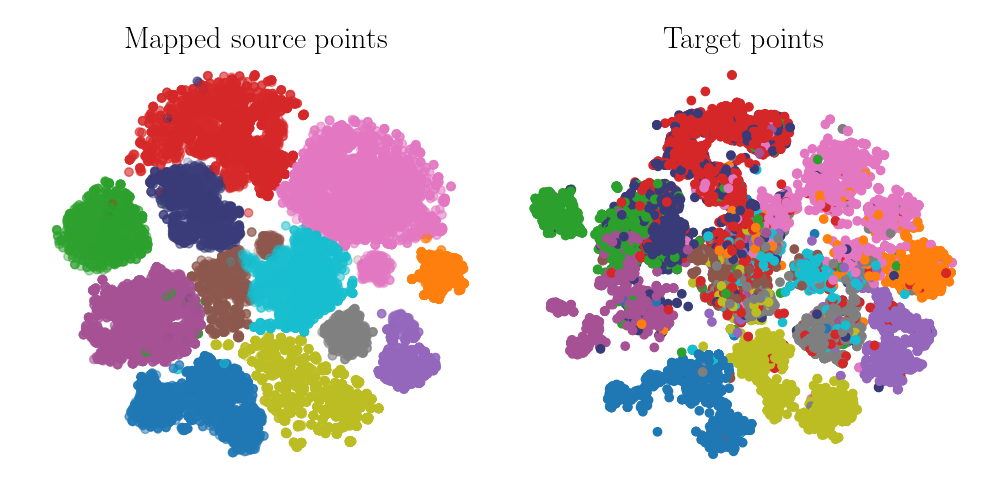}
        \caption{GMM-OTDA$_{T}$}
    \end{subfigure}
    \caption{\textcolor{black}{t-\gls{sne} visualization \gls{ot} map-based \gls{da} techniques on the VisDA-C benchmark with ViT-16-b features. Colors represent different classes. In (d), stars represent the components of source and target \glspl{gmm}, obtained through the conditional \gls{em}, and the \gls{em} algorithm respectively.}}
    \label{fig:visda_tsne}
\end{figure}

\textcolor{black}{Finally, we analyze how the different mapping strategies match source and target domain measures. We summarize this analysis in Figure~\ref{fig:visda_tsne}, where we show the t-\gls{sne} of the concatenation of mapped source and target domain data. In contrast with OTDA$_{EMD}$, OTDA$_{Sinkhorn}$ and GMM-OTDA$_{T}$, OTDA$_{affine}$ does not manage to match source and target data, mainly due the simplicity of the Gaussian assumption. Furthermore, GMM-OTDA$_{T}$ manages to map source domain data in a way that does not mixes the classes (for instance, compare Figure~\ref{fig:visda_tsne} (a) with (e)). The label propagation approach is also discriminative of target domain classes, as is evidenced in Figure~\ref{fig:visda_tsne} (d). These considerations explain the superior performance of GMM-OTDA$_{T}$ with ViT-16-b features.}

\newpage

\subsection{Cross-Domain Fault Diagnosis}\label{sec:cdfd}

For \gls{cdfd} benchmarks, we follow the experimental settings of~\cite{montesuma2023dadil},~\cite{montesuma2022cross} and~\cite{montesuma2024benchmarking}, which roughly follows a similar idea to visual adaptation tasks. This means that we pre-train a neural network with source domain data, then use its encoder for feature extraction. We refer readers to the original papers, and our appendix, for further information.

\noindent\textbf{Case Western Reserve University Benchmark.} With respect other benchmarks, the \gls{cwru} has the most number of samples per domain, i.e., 8000. In this case, InfoOT was intractable due its computational complexity. Furthermore, this benchmark illustrate the advantage of employing a grouping technique for enforcing the class structure in \gls{ot}.

\begin{table}[ht]
    \centering
    \resizebox{\linewidth}{!}{\begin{tabular}{cccccccccc}
        \toprule
        Task & Baseline & OTDA$_{EMD}$ & OTDA$_{Sink}$ & OTDA$_{affine}$ & InfoOT$_{b}$ & InfoOT$_{c}$ & HOT-DA & GMM-OTDA$_{MAP}$ & GMM-OTDA$_{T}$\\
        \midrule
        $A \rightarrow B$  & 51.12 & 72.00 & 75.19  & 78.12 & - & - & 69.88  & \textbf{79.75} & \textbf{79.75}\\
        $A \rightarrow C$  & 62.88 & 94.12 & \textbf{100.00} & 95.62 & - & - & \textbf{100.00} & 99.94 & \textbf{100.00}\\
        $B \rightarrow A$  & 42.50 & 76.12 & 78.50  & 75.88 & - & - & 79.75  & \textbf{80.00} & \textbf{80.00}\\
        $B \rightarrow C$  & 37.44 & 77.62 & 78.88  & 75.38 & - & - & 79.81  & 79.56 & \textbf{79.94}\\
        $C \rightarrow A$  & 52.81 & 98.38 & 99.25  & 94.12 & - & - & 98.75  & 99.12 & \textbf{99.88}\\
        $C \rightarrow B$  & 55.62 & 70.25 & 74.50  & 75.50 & - & - & 83.12  & 79.75 & \textbf{80.00}\\
        Avg.               & 50.40 & 81.42 & 84.39  & 82.44 & - & - & 85.22  & 86.35 & \textbf{86.59}\\
        \bottomrule
    \end{tabular}}
    \caption{Experimental results on the CWRU benchmark. For each task (i.e., each row), we highlight the best performing method in bold. Overall, InfoOT did not have a reasonable running time due the large number of samples on each domain.}
    \label{tab:cwru_results}
\end{table}

\vspace{5mm}

\noindent\textbf{Continuous Stirred Tank Reactor.} As covered in~\cite{montesuma2022cross}, the \gls{cstr} process carries an exothermic reaction $A \rightarrow B$. The reactor is jacketed, and an inflow of coolant is controlled by a \gls{pid} controller as described in~\cite{pilario2017process}. From this reactor, a set of 7 variables are measured throughout simulation, corresponding to different temperatures, concentrations and flow-rates. We refer readers to~\cite{montesuma2022cross} for further details. Associated with this process, there are a set of 12 different faults, ranging from process and sensors faults, and input disturbances. On top of these 12 faults, there is the no-fault scenario, characterizing a classification problem with 13 classes.

\begin{table*}[ht]
\centering
        \begin{tabular}{lccccccc}
            \toprule
            Target Domain & 1 & 2 & 3 & 4 & 5 & 6 &\multirow{3}{*}{Score}\\
            Reaction Order $(N)$ & 1.0 & 1.0 & 1.0 & 0.5 & 1.5 & 2.0 & \\
            Parameter Noise $(\eta)$ & $10\%$ & $15\%$ &  $20\%$ & $15\%$ & $15\%$ & $15\%$ & \\
            \midrule
            Baseline & 69.23 & 67.30 & \textbf{73.07} & 53.84 & 63.46 & \textbf{57.69} & 64.10\\
            OTDA$_{EMD}$ & 71.15 & 71.15 & 71.15 & 61.53 & 57.69 & 50.00 & 63.78\\
            OTDA$_{Sink}$ & 67.31 & 67.31 & 69.23 & 55.76 & 51.92 & 53.84 & 60.89\\
            OTDA$_{Affine}$ & 65.38 & 71.15 & 71.15 & 61.54 & \textbf{65.38} & 53.84 & 64.74\\
            InfoOT$_{b}$ & 67.31 & 67.31 & 67.31 & 50.00 & 51.92 & 40.07 & 58.65\\
            InfoOT$_{c}$ & 71.15 & 67.31 & 71.15 & 53.84 & 51.92 & 48.07 & 60.57\\
            HOT-DA & 55.77 & 40.38 & 44.23 & 55.77 & 40.38 & 42.31 & 46.47\\
            GMM-OTDA$_{MAP}$ & \textbf{78.84} & \textbf{73.07} & \textbf{73.07} & 61.54 & 57.69 & 55.77 & \textbf{66.67}\\
            GMM-OTDA$_{T}$ & 76.92 & \textbf{73.07} & \textbf{73.07} & \textbf{63.46} & 53.84 & 50.00 & 65.04\\
            \bottomrule
        \end{tabular}
    \vskip0.1cm
    \caption{Average classification accuracy with confidence intervals over a 5-fold cross-validation.}
    \label{tab:cstr_results}
\end{table*}

The different domains in this benchmark correspond to changes in the process conditions. These are of 2 kinds. First, one introduces a noise, $\eta$, in the process parameters (e.g., reactor or jacket volume), reflecting the possible uncertainty in the mathematical modeling of the reactor. Second, one changes the reaction order, $N$, of the reaction $A \rightarrow B$, which drastically changes the dynamics of the system. As~\cite{montesuma2022cross}, we consider $\eta \in \{0.1, 0.15, 0.2\}$, and $N \in \{1, 0.5, 1.5, 2\}$. The source domain corresponds to $N = 1$, $\eta = 0.0$, whereas the 6 different targets correspond to combinations of $\eta$ and $N$. These are shown in Table~\ref{tab:cstr_results}.

The \gls{cstr} benchmark poses a few challenges. First, it has a small number of samples on domains. While the source is composed of 1300 samples, each target only has 260. Second, each sample lies in a $1400-$dimensional space. Third, target domains are noisy, due the parameter noise $\eta$. As a result, most methods have difficulty in adapting, and performance usually degrades for more intense shifts (e.g., $N=2.0$ and $\eta = 0.15$). However, GMM-OTDA$_{MAP}$ and GMM-OTDA$_{T}$ outperform other methods.

\begin{figure}[ht]
    \begin{subfigure}{0.23\linewidth}
        \includegraphics[width=\linewidth]{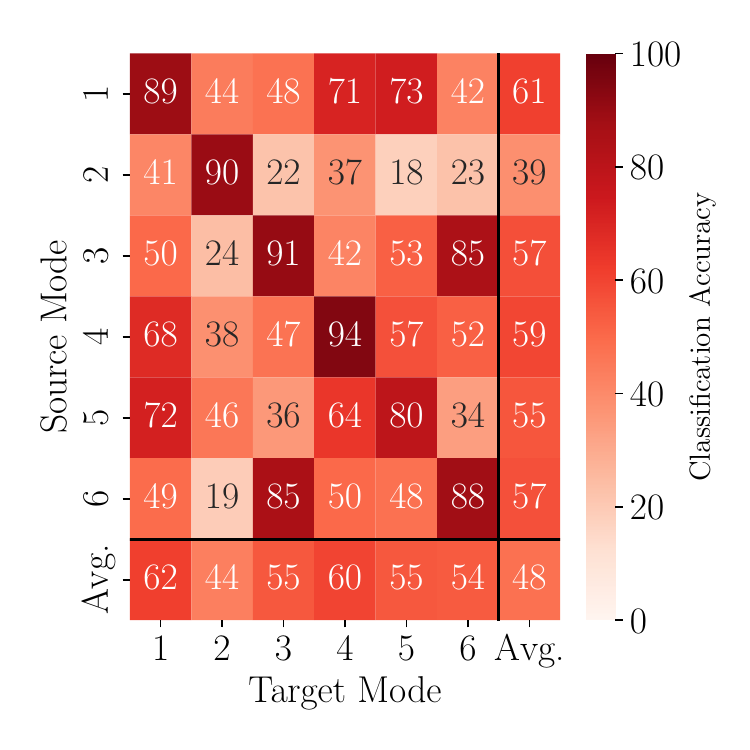}
        \caption{Baseline}
    \end{subfigure}\hspace{2mm}
    \begin{subfigure}{0.23\linewidth}
        \includegraphics[width=\linewidth]{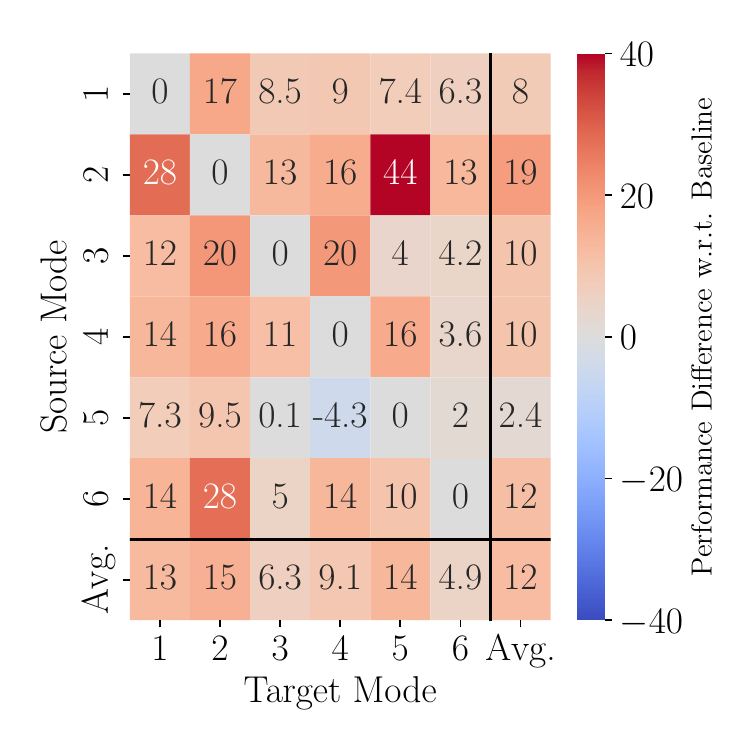}
        \caption{OTDA$_{EMD}$}
    \end{subfigure}\hspace{2mm}
    \begin{subfigure}{0.23\linewidth}
        \includegraphics[width=\linewidth]{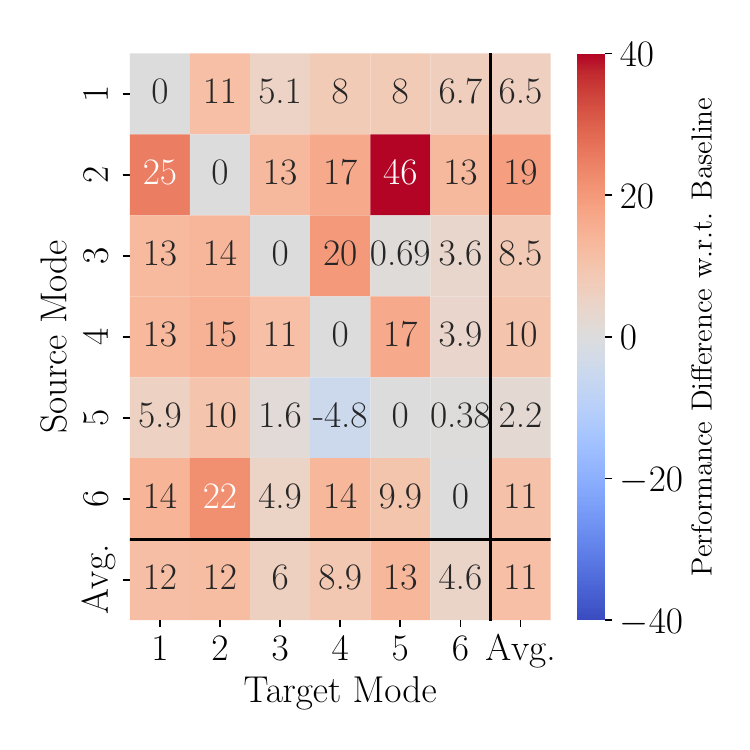}
        \caption{OTDA$_{Sinkhorn}$}
    \end{subfigure}\hspace{2mm}
    \begin{subfigure}{0.23\linewidth}
        \includegraphics[width=\linewidth]{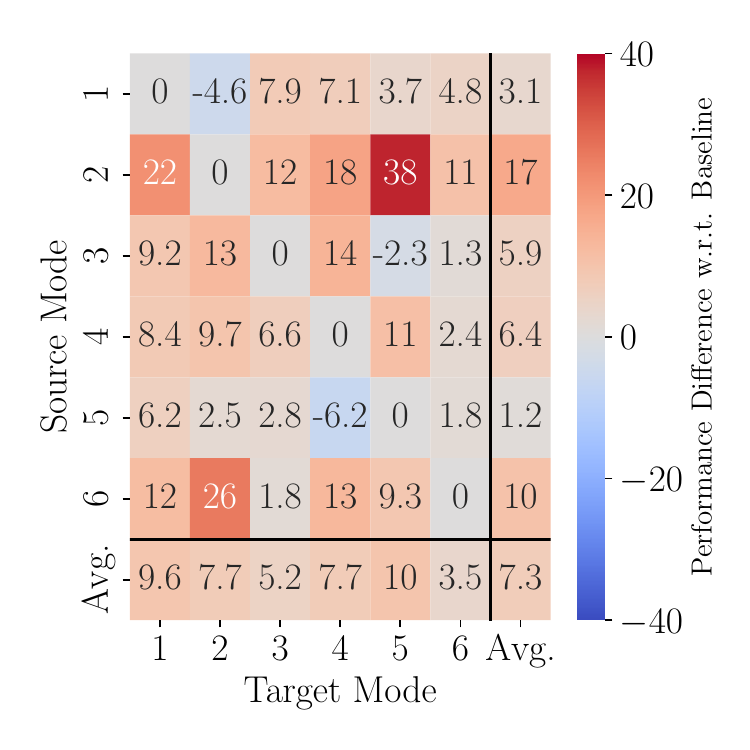}
        \caption{OTDA$_{Affine}$}
    \end{subfigure}\\
    \begin{subfigure}{0.23\linewidth}
        \includegraphics[width=\linewidth]{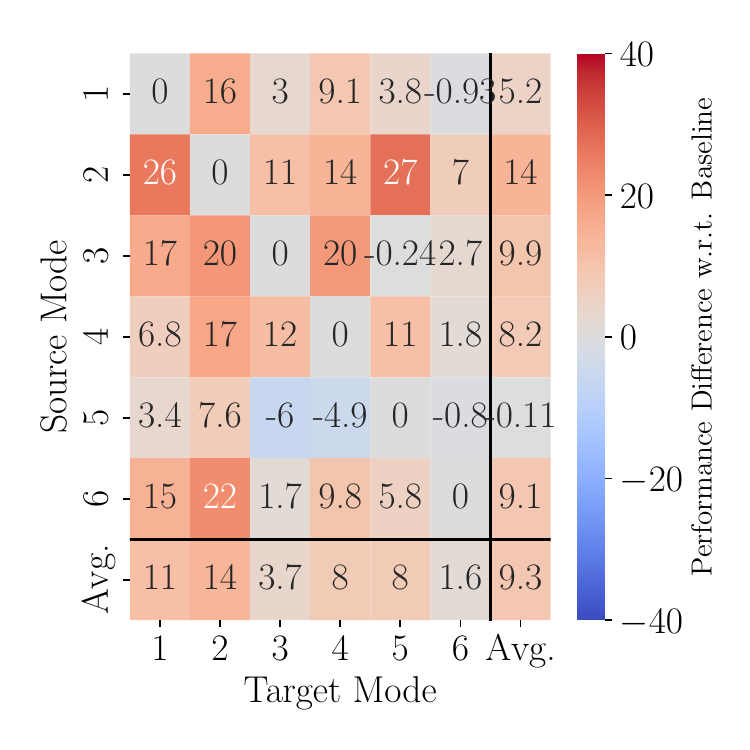}
        \caption{HOTDA}
    \end{subfigure}
    \begin{subfigure}{0.23\linewidth}
        \includegraphics[width=\linewidth]{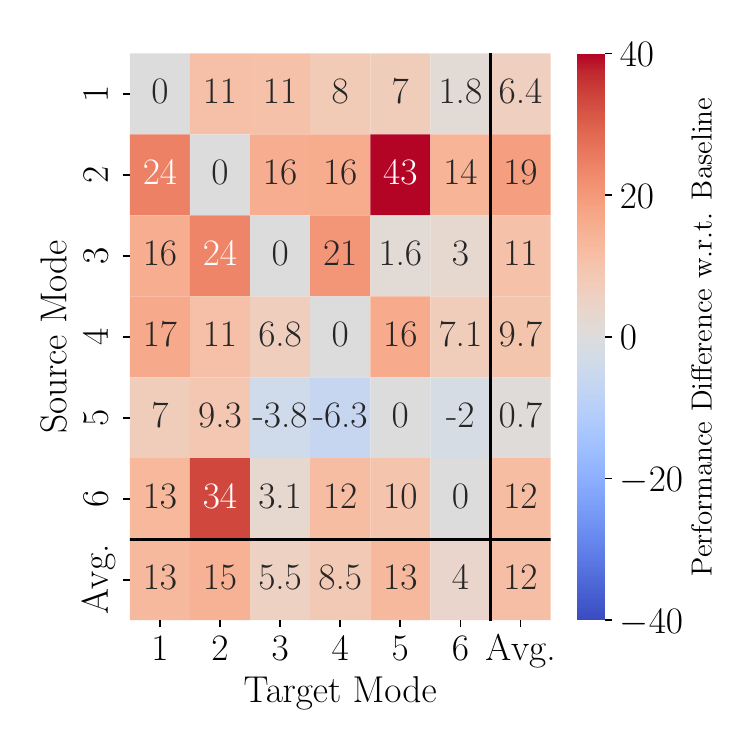}
        \caption{GMM-OTDA$_{M}$}
    \end{subfigure}\hspace{2mm}
    \begin{subfigure}{0.23\linewidth}
        \includegraphics[width=\linewidth]{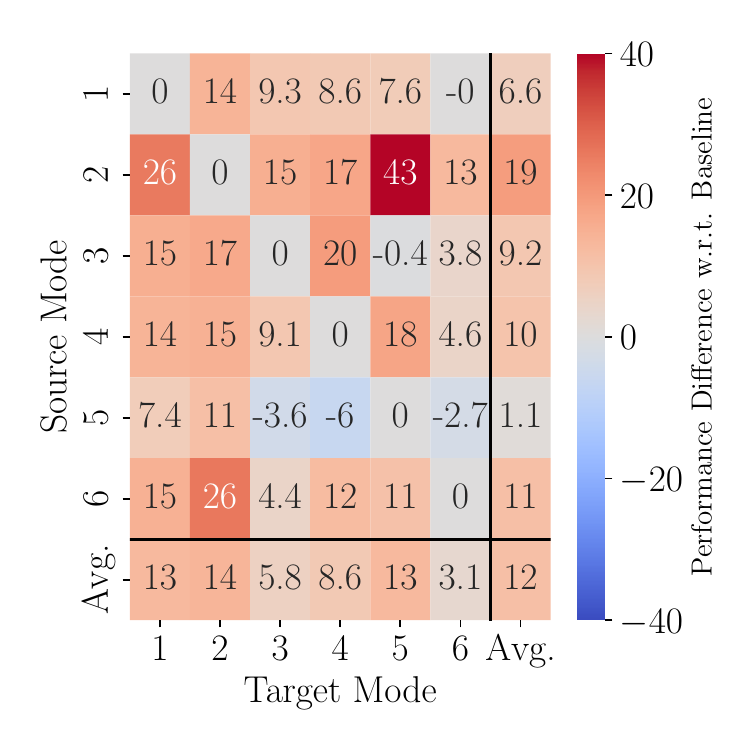}
        \caption{GMM-OTDA$_{T}$}
    \end{subfigure}\hspace{2mm}
    \begin{subfigure}{0.23\linewidth}
        \includegraphics[width=\linewidth]{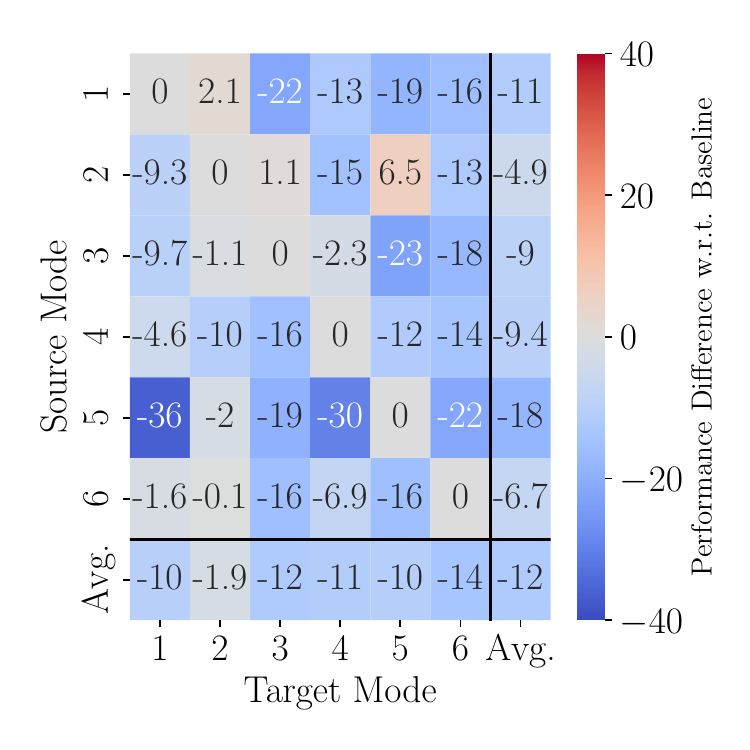}
        \caption{DANN}
    \end{subfigure}\\
    \begin{subfigure}{0.23\linewidth}
        \includegraphics[width=\linewidth]{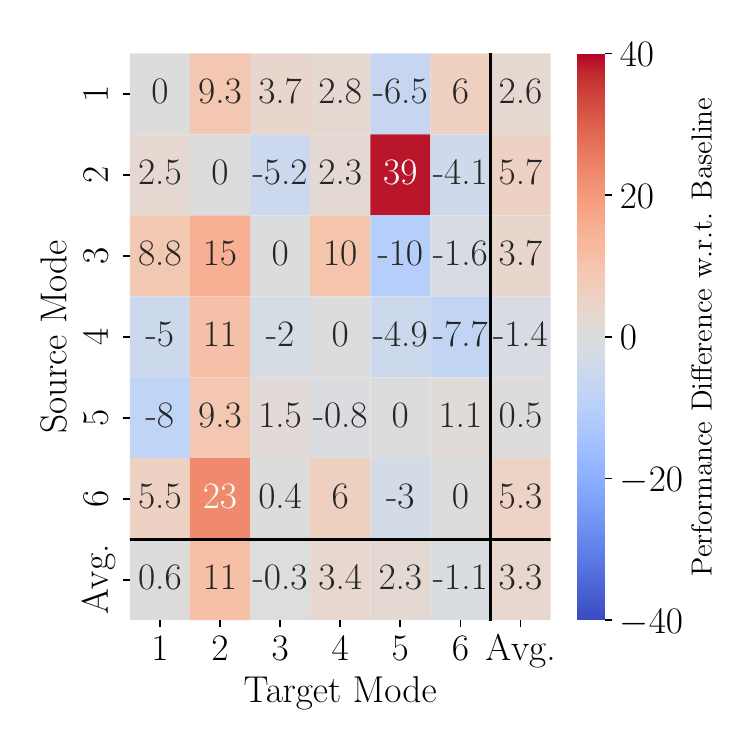}
        \caption{DAN}
    \end{subfigure}
    \begin{subfigure}{0.23\linewidth}
        \includegraphics[width=\linewidth]{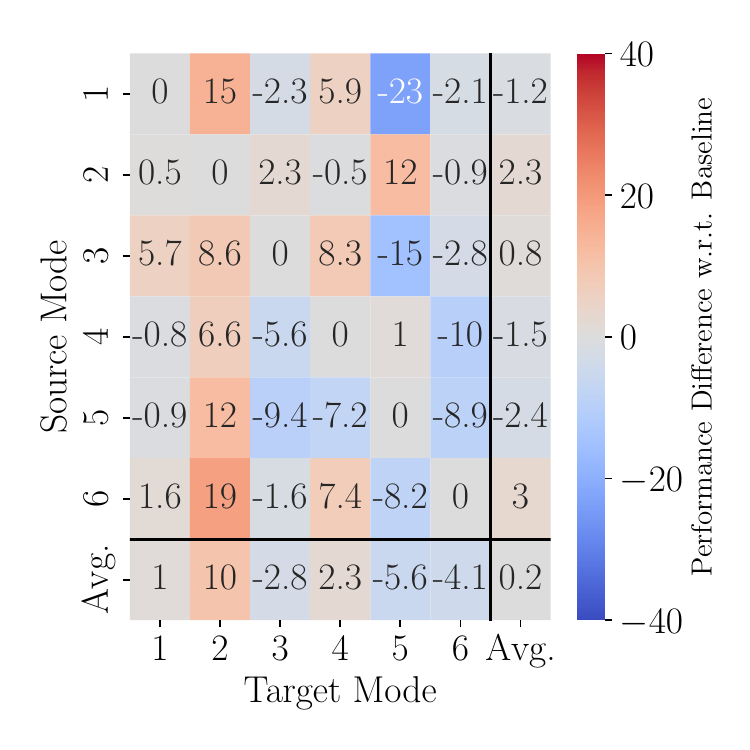}
        \caption{WDGRL}
    \end{subfigure}\hspace{2mm}
    \begin{subfigure}{0.23\linewidth}
        \includegraphics[width=\linewidth]{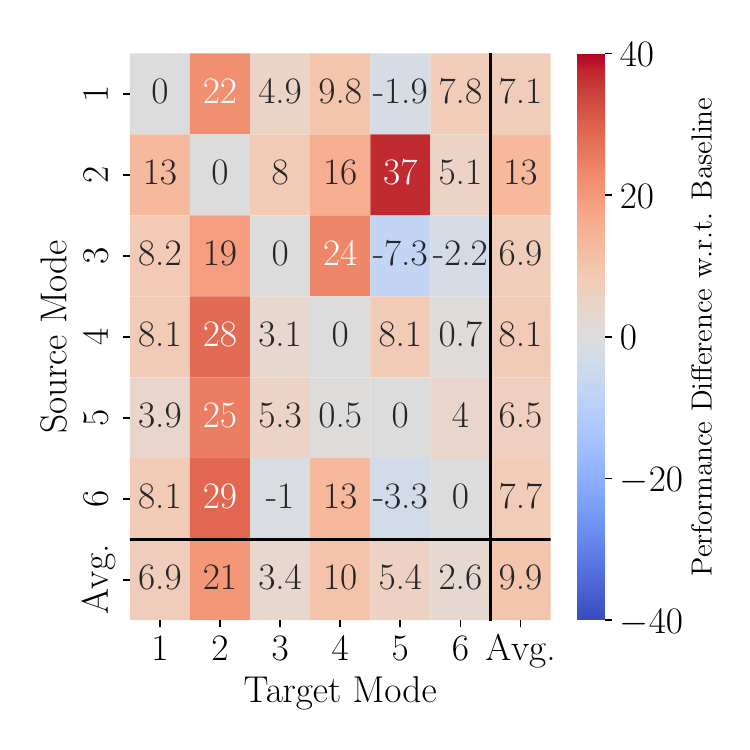}
        \caption{DeepJDOT}
    \end{subfigure}\hspace{2mm}
    \caption{Domain adaptation results on the Tennessee Eastman Process. In (a), we show the baseline adaptation tasks, where each row represents a source domain, and each column represents a target domain. From (b) to (k), we show the performance offset with respect (a) of adaptation algorithms. Note that (i) through (k) are deep learning-based algorithms.}
    \label{fig:tep_ssda}
\end{figure}

\noindent\textbf{Tennessee Eastman Process.} Our last experiment consists of the \gls{te} process, a benchmark widely used by the chemical engineering community~\cite{reinartz2021extended}. This benchmark has the largest number of domains, i.e., $6$. Each of these domains is characterized by a different mode of production for the products of a chemical reaction, which affects the measured signals from the chemical plant. We summarize our results in Figure~\ref{fig:tep_ssda}, which comprises the $30$ adaptation tasks.

\textcolor{black}{For this benchmark, we compare 11 methods. We divide those into \emph{shallow} \gls{da} methods, and \emph{deep} \gls{da} methods. Shallow methods try to cope with distributional shift by transforming or re-weighting the samples in a feature space. In the case of this benchmark, we use the encoder's activations as features. In contrast, deep methods cope with distribution shift by learning discriminative, \emph{domain invariant} features, by penalizing the encoder's parameters $\theta_{g}$ so that, after encoding the data points, the domains are indistinguishable from each other. Besides the 6 shallow methods compared throughout this paper, we also consider classic deep methods, such as \gls{dann}~\cite{ganin2016domain}, \gls{dan}~\cite{ghifary2014domain}, \gls{wdgrl}~\cite{shen2018wasserstein} and \gls{deepjdot}~\cite{damodaran2018deepjdot}.}

In comparison with other benchmarks, the \gls{te} feature vectors have fewer dimensions (i.e., $128$). As a result, OTDA$_{EMD}$ is the best performing method. However, GMM-OTDA manages to improve over OTDA$_{affine}$ and HOT-DA. Overall, our methods are especially better on harder adaptation tasks, such as $6 \rightarrow 2$ and $3 \rightarrow 2$. We refer readers to the exploratory data analysis of~\cite{montesuma2024benchmarking} for further insights on why these adaptation tasks are harder.

\textcolor{black}{Furthermore, note that shallow methods are comparatively better to deep methods. Indeed, the deep neural nets use considerably less data than, for instance, the image benchmarks considered in our experiments section. In this latter case, previously to the fine-tuning step on the source domain data, ResNets and ViTs are pre-trained on the ImageNet benchmark~\cite{deng2009imagenet}, which provides a good starting model for natural image classification. In the context of the \gls{te} process benchmark, this is not possible since data are time series of a specific chemical process.}

\subsection{Ablations and Visualization}\label{sec:ablations}

\begin{figure}[ht]
    \centering
    \includegraphics[width=0.7\linewidth]{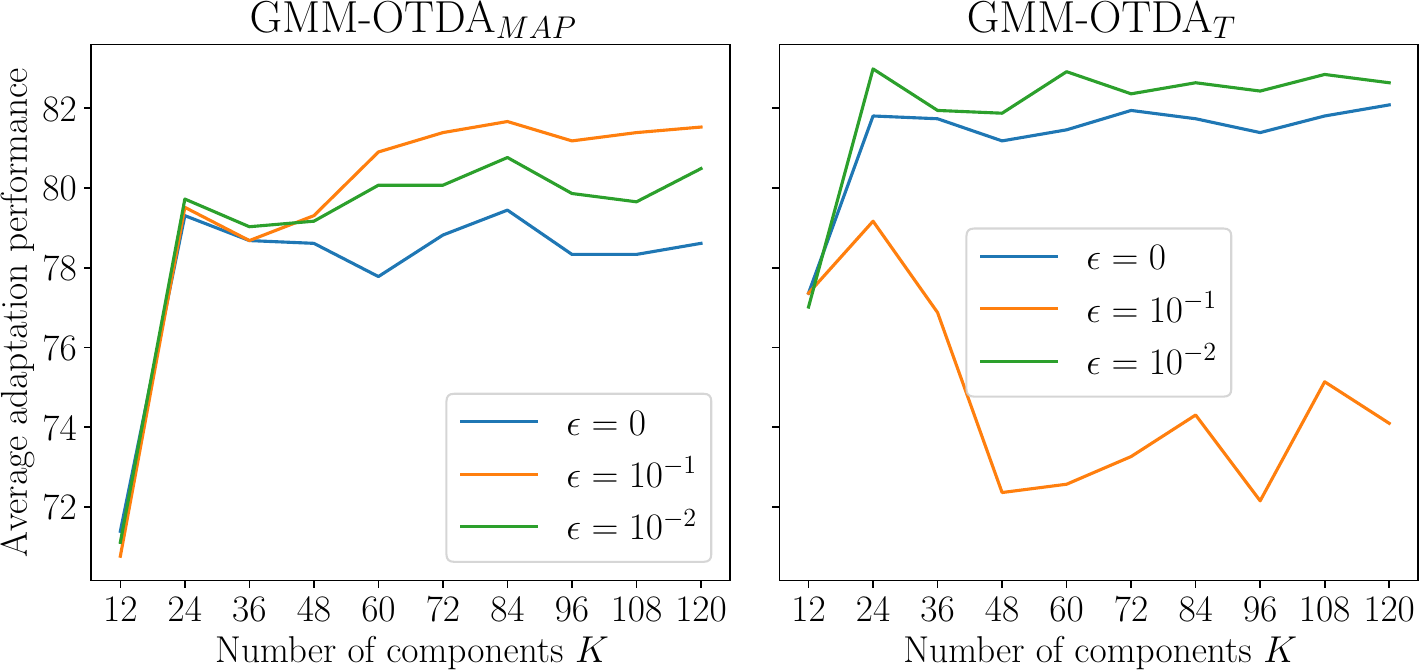}
    \caption{Ablation on number of components $K$, and entropic penalty $\epsilon$, for the \gls{map} estimation strategy based on labeled propagation (left), and the mapping estimation strategy (right).}
    \label{fig:components_reg_e}
\end{figure}

\noindent\textbf{Ablating the number of components and entropic regularization.} In this experiment, we ablate the two parameters of our methods, namely, the number of components $K$, and the entropic regularization $\epsilon$. Recall that we normalize the ground-cost by the maximum value, i.e., $\tilde{C}_{ij} = C_{ij}/(\text{max}_{ij}C_{ij})$, which improves the numerical stability of the Sinkhorn algorithm. We summarize our results in Figure~\ref{fig:components_reg_e}.

For the number of components $K$, the relationship with performance is mostly clear. Indeed, except for $\epsilon = 10^{-1}$ on the mapping strategy, \emph{using more components enhances performance}. Note that, even though this implies a more complex \glspl{gmm}, we still have far less components than samples ($n=600$ per domain, i.e., $5$ times more samples than components).

For the entropic penalty $\epsilon$, we have two drastically different scenarios. For the \gls{map} estimation, using higher entropic regularization coefficients improve performance, whereas the mapping strategy works better for smaller regularization coefficients (or exact \gls{ot}). While this may seem contradictory, we note that, in the mapping strategy, we actually filter out irrelevant matchings between components based on a parameter $\tau$. However, the entropic regularization is known to generate smoother couplings, which means that more entries of $\omega$ are non-zero, or possibly greater than a fixed $\tau$. As a consequence, the mapping strategy ends up behaving like $T_{rand}$, which causes a bad reconstruction for the target domain. Naturally, this effect gets amplified with more components in both \glspl{gmm}, as there are more possible matchings.

\begin{figure}[ht]
    \centering
    \includegraphics[width=0.7\linewidth]{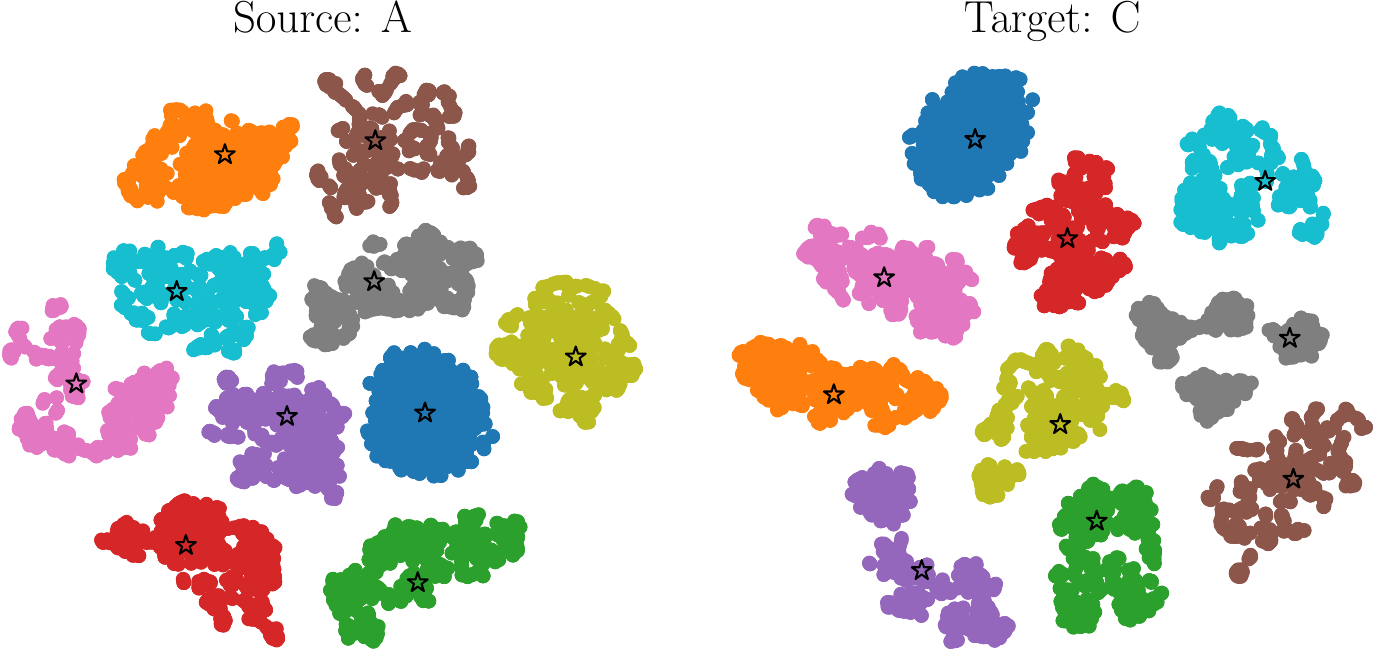}
    \caption{Source and target domain samples alongside the centroids (denoted by stars) found through \gls{em}. colors reflect the different classes.}
    \label{fig:centroids-cwru}
\end{figure}

\begin{figure}[ht]
    \centering
    \begin{subfigure}{0.24\linewidth}
        \includegraphics[width=\linewidth]{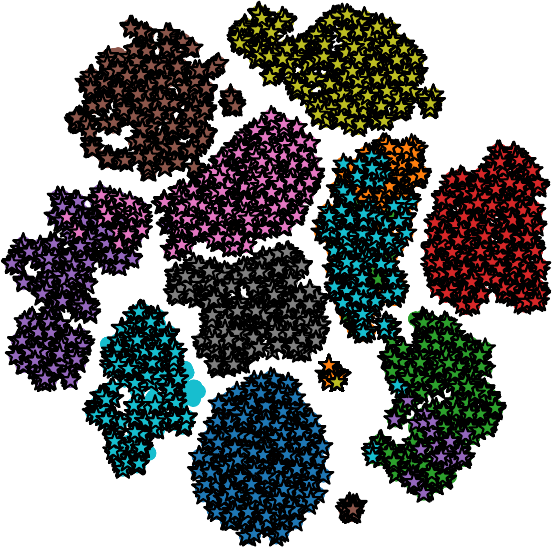}
        \caption{OTDA$_{EMD}$}
    \end{subfigure}\hfill
    \begin{subfigure}{0.24\linewidth}
        \includegraphics[width=\linewidth]{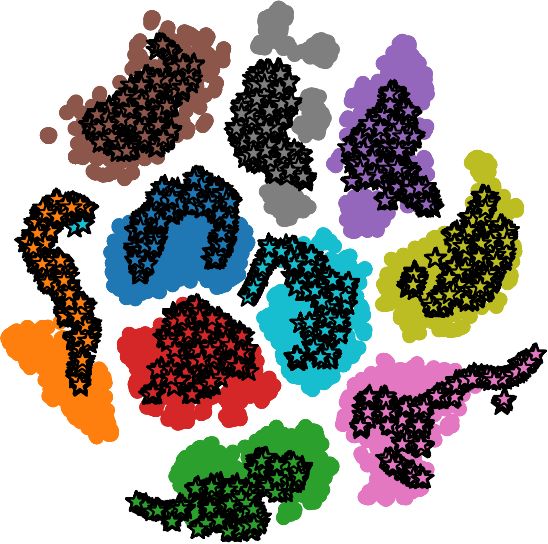}
        \caption{OTDA$_{Sink}$}
    \end{subfigure}\hfill
    \begin{subfigure}{0.24\linewidth}
        \includegraphics[width=\linewidth]{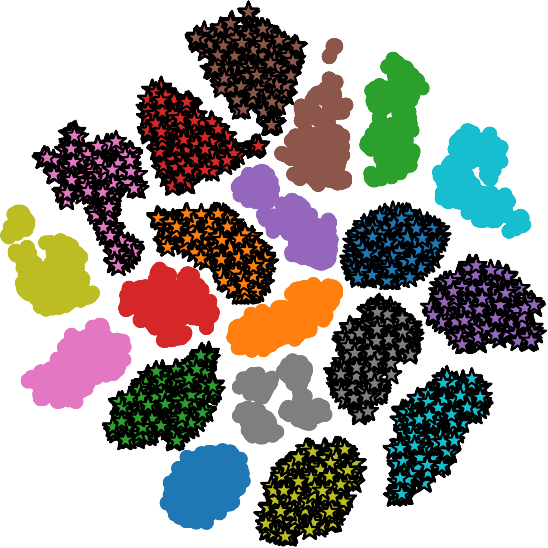}
        \caption{OTDA$_{affine}$}
    \end{subfigure}\hfill
    \begin{subfigure}{0.24\linewidth}
        \includegraphics[width=\linewidth]{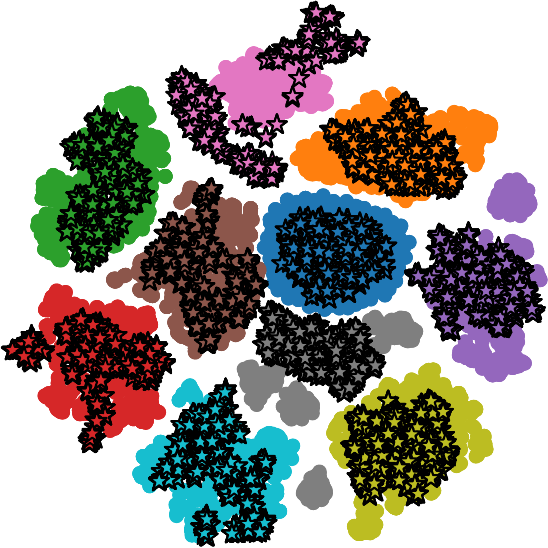}
        \caption{GMM-OTDA$_{T}$}
    \end{subfigure}
    \caption{t-\gls{sne} visualisation of mapped samples to the target domain, on the CWRU task $A \rightarrow C$.}
    \label{fig:vis_tsne}
\end{figure}

\noindent\textbf{Visualizing components and mapped samples.} In this experiment, we use the \gls{cwru} benchmark adaptation task $A \rightarrow C$. We start by embedding the source and target domain data with the t-\gls{sne} technique of~\cite{van2008visualizing}. We do this in 2 separate plots, where we concatenate the source domain features with the centroids obtained by running the \gls{em} algorithm (resp. target). This visualization is shown in Figure~\ref{fig:centroids-cwru}. Next, we map samples from the source to the target domain, with the various strategies described in this section, with the exception of InfoOT, which did not had a reasonable running time. These are shown in Figure~\ref{fig:vis_tsne}.

Overall, while the exact \gls{ot} solution provides a measure that better reflects the feature positions, it mixes the classes, as evidenced by the orange and blue classes being mapped to the same place, as well as the green and violet classes. This phenomenon does not happen for other methods, at the cost of having mapped points distributed in a different way. However, for OTDA$_{sink}$ and GMM-OTDA$_{T}$, the mapped points better respect the class boundaries. For OTDA$_{affine}$, note that the mapped distribution does not actually match the target. Overall, we achieve a better mapping through the \gls{gmm} modeling.

\subsection{Running Time Analysis}\label{sec:runtime}

\textcolor{black}{Besides our remark~\ref{remark:complexity}, we also run a running time analysis of the tested algorithms. Our experimental setting is as follows. We use the adaptation task $A \rightarrow W$ of the Office 31 benchmark. In this case, $n_{S} = 2817$, $n_{T} = 624$, $n_{c} = 31$ and $d= 2048$. We ran each algorithm 10 independent times, then computed the mean and standard deviation of their running time. Our results are reported on tables~\ref{tab:algorithm_comparison} and~\ref{tab:components_performance}.}
 
\begin{table}[ht]
\centering
\begin{tabular}{lcc}
\toprule
Algorithm & Running Time (seconds) & Accuracy (\%) \\ 
\midrule
OTDA$_{\text{EMD}}$ & 0.775 $\pm$ 0.007 & 74.27 \\ 
OTDA$_{\text{Sink}}$ & 14.119 $\pm$ 0.125 & 80.12 \\ 
OTDA$_{\text{Affine}}$ & 8.503 $\pm$ 0.097 & 80.12 \\ 
InfoOT & 105.407 $\pm$ 0.716 & 79.53 \\ 
HOTDA & 2.508 $\pm$ 0.023 & 73.68 \\ 
\bottomrule
\end{tabular}
\caption{\textcolor{black}{Running time (in seconds) and classification accuracy (in \%) of different \gls{ot}-based strategies.}}
\label{tab:algorithm_comparison}
\end{table}

\textcolor{black}{Starting from table~\ref{tab:algorithm_comparison}, the fastest algorithm is OTDA$_{\text{EMD}}$, which has complexity $\mathcal{O}(n^{3}\log n)$. In comparison, OTDA$_{\text{Sink}}$ has complexity $\mathcal{O}(n^{2})$ per iteration. Here, it is important to note that we run the Sinkhorn algorithm until convergence, for $1000$ iterations, which explains its superior running time. It is noteworthy that OTDA$_{affine}$ also has a higher running time, since its complexity is dimension-dependent, i.e., $\mathcal{O}(d^{3})$. Due the high dimensional character of the data at hand, this results in a higher running time.}

\textcolor{black}{Another example of higher running time comes from HOTDA, which solves $n_{c}^{2} - n_{c} = 465$ sub empirical \gls{ot} problems, resulting in a higher running time in comparison with OTDA$_{\text{EMD}}$. Finally, it is noteworthy that InfoOT is considerably slower than other methods, due to its $\mathcal{O}(n^{3})$ complexity \emph{by iteration}.}

\textcolor{black}{In comparison with previous methods, we show the running time and classification accuracy of GMM-OTDA$_{\text{T}}$ and GMM-OTDA$_{\text{MAP}}$ for $K=\{31, 62, \cdots, 217\}$. We do so for $\epsilon = 10^{-2}$, which in practice yielded the best empirical performance. As a result, the running time should be directly compared to OTDA$_{\text{Sink}}$. For all number of components, our algorithm has an inferior running time to almost all methods, with the expection of OTDA$_{\text{EMD}}$ and HOTDA.}

\begin{table}[ht]
\centering
\begin{tabular}{cccc}
\toprule
\text{Number of Components} & \text{Running Time (seconds)} & GMM-OTDA$_{\text{T}}$ & GMM-OTDA$_{\text{MAP}}$\\ 
\midrule
31 & 0.742 $\pm$ 0.074 & 78.94 & 64.91 \\ 
62 & 1.329 $\pm$ 0.015 & 79.53 & 70.76 \\ 
93 & 1.995 $\pm$ 0.035 & 76.02 & 70.17 \\ 
124 & 2.679 $\pm$ 0.051 & 80.70 & 74.85 \\ 
155 & 3.517 $\pm$ 0.025 & 78.36 & 75.44 \\ 
186 & 4.343 $\pm$ 0.090 & 80.70 & 70.17 \\ 
217 & 5.069 $\pm$ 0.059 & 77.77 & 76.02 \\ 
\bottomrule
\end{tabular}
\caption{\textcolor{black}{Running time (in seconds) and classification accuracy (in \%) of GMM-OTDA$_{\text{T}}$ and GMM-OTDA$_{\text{MAP}}$ as a function of number of components.}}
\label{tab:components_performance}
\end{table}
\section{Conclusion}\label{sec:conclusion}

In this paper, we consider the \gls{gmmot} framework of~\cite{delon2020wasserstein} as a candidate for \gls{uda}. Based on probability and \gls{ot} theory, we devise 2 new effective strategies for \gls{uda}. The label propagation interprets the \gls{ot} plan between \gls{gmm} components as a joint probability distribution over source-target component pairs. This modeling choice allows us to predict the label of target \gls{gmm} components through a label propagation equation similar to~\cite{redko2019optimal}. Furthermore, we propose a mapping strategy that transports samples from the same component together through an affine map, which has 2 advantages. First, it enforces group sparsity~\cite{courty2016otda}. Second, it has an analytical form in terms of \gls{gmm} parameters. We show through a series of 85 \gls{uda} tasks that our methods outperform, or are competitive with the state-of-the-art in shallow domain adaptation, while being scalable with both number of samples $n$, and number of dimensions $d$. Our work further confirms previous studies on the intersection of \glspl{gmm} and \gls{uda}, such as~\cite{montesuma2024lighter}, showing that the \gls{gmmot} is a powerful candidate for shallow domain adaptation.

\bibliographystyle{unsrt}
\bibliography{ref.bib}

\appendix
\newpage

\section{Additional Details about Benchmarks}

\begin{table}[ht]
    \centering
    \begin{minipage}[b]{.48\linewidth}
        \centering
        \resizebox{\linewidth}{!}{%
        \begin{tabular}{ccccc}
            \toprule
            Benchmark                          & Domains         & Backbone                    & \# Samples & \# Classes           \\
            \midrule
            \multirow{4}{*}{ImageCLEF}         & Caltech (C)     & \multirow{4}{*}{ResNet 50}  & 600        & \multirow{4}{*}{12}  \\
                                               & Bing (B)        &                             & 600        &                      \\
                                               & ImageNet (I)    &                             & 600        &                      \\
                                               & Pascal (P)      &                             & 600        &                      \\
                                               & Total           &                             & 2400       &                      \\
            \midrule
            \multirow{4}{*}{Caltech-Office 10} & Amazon (A)      & \multirow{4}{*}{ResNet 101} & 958        & \multirow{4}{*}{10}  \\
                                               & dSLR (D)        &                             & 157        &                      \\
                                               & Webcam (W)      &                             & 295        &                      \\
                                               & Caltech (C)     &                             & 1123       &                      \\
                                               & Total           &                             & 2533       &                      \\
            \midrule
            \multirow{3}{*}{Office 31}         & Amazon (A)      & \multirow{3}{*}{ResNet 50}  & 2817       & \multirow{3}{*}{31}  \\
                                               & dSLR (D)        &                             & 498        &                      \\
                                               & Webcam (W)      &                             & 795        &                      \\
                                               & Total           &                             & 4110       &                      \\
            \midrule
            \multirow{4}{*}{Office-Home}       & Art (Ar)        & \multirow{4}{*}{ResNet 101} & 2427       & \multirow{4}{*}{65}  \\
                                               & Clipart (Cl)    &                             & 4365       &                      \\
                                               & Product (Pr)    &                             & 4439       &                      \\
                                               & Real World (Rw) &                             & 4357       &                      \\
                                               & Total           &                             & 15588      &                      \\
            \bottomrule
        \end{tabular}
        }
        \subcaption{Visual Domain Adaptation Benchmarks}
    \end{minipage}\hfill
    \begin{minipage}[b]{.48\linewidth}
        \centering
        \resizebox{\linewidth}{!}{%
        \begin{tabular}{ccccc}
            \toprule
            Benchmark                          & Domains         & Backbone                    & \# Samples & \# Classes           \\
            \midrule
            \multirow{4}{*}{CWRU}              & 1772rpm (A)     & \multirow{4}{*}{MLP}        & 8000       & \multirow{4}{*}{10}  \\
                                               & 1750rpm (B)     &                             & 8000       &                      \\
                                               & 1730rpm (C)     &                             & 8000       &                      \\
                                               & Total           &                             & 24000      &                      \\
            \midrule
            \multirow{6}{*}{TEP}              & Mode 1          & \multirow{6}{*}{Fully Convolutional} & 2900      & \multirow{6}{*}{29}  \\
                                               & Mode 2          &                             & 2845       &                      \\
                                               & Mode 3          &                             & 2899       &                      \\
                                               & Mode 4          &                             & 2865       &                      \\
                                               & Mode 5          &                             & 2883       &                      \\
                                               & Mode 6          &                             & 2897       &                      \\
                                               & Total           &                             & 17289      &                      \\
            \midrule
            \multirow{7}{*}{CSTR}              & $N=1.0, \epsilon=0.00$ & \multirow{7}{*}{-}   & 1300       & \multirow{7}{*}{13}  \\
                                               & $N=1.0, \epsilon=0.10$ &                             & 260        &                      \\
                                               & $N=1.0, \epsilon=0.15$ &                             & 260        &                      \\
                                               & $N=0.5, \epsilon=0.15$ &                             & 260        &                      \\
                                               & $N=1.5, \epsilon=0.15$ &                             & 260        &                      \\
                                               & $N=2.0, \epsilon=0.15$ &                             & 260        &                      \\
                                               & Total           &                             & 2860       &                      \\
            \bottomrule
        \end{tabular}
        }
        \subcaption{Cross-Domain Fault Diagnosis Benchmarks}
    \end{minipage}
    \caption{Overview of Visual Domain Adaptation and Cross-Domain Fault Diagnosis benchmarks}
    \label{tab:overview_combined_benchmarks}
\end{table}

In table~\ref{tab:overview_combined_benchmarks}, we show an overview of the used benchmarks. We run our experiments on 4 visual \gls{da} datasets, and 3 \gls{cdfd} datasets. For vision, we use \glspl{resnet} pre-trained on ImageNET as the backbone. For each adaptation task (e.g., $C \rightarrow B$ in ImageCLEF) we fine tune the network using labeled source domain data. For all methods, we extract the features of source, and target domain, using the fine-tuned checkpoint. The size of the \gls{resnet} is used to agree with previous research, such as~\cite{peng2019moment} and~\cite{montesuma2023dadil}. We refer readers to\footnote{\faGithub\quad\url{https://github.com/eddardd/DA-baselines}} for further technical details on the fine-tuning of vision backbones.

For \gls{cdfd}, we considered the same setting as previous works using these benchmarks, such as~\cite{montesuma2023dadil},~\cite{montesuma2024benchmarking} and~\cite{montesuma2022cross}, for \gls{cwru}, \gls{te} process and \gls{cstr} respectively. For \gls{cwru}, we extract windows out of raw signals of size 2048, then get the frequency representation for these windows using a fast Fourier transform. These are treated as $2048$ feature vectors, that are then fed to a neural network. In the \gls{te} process, we consider the same setting of~\cite{montesuma2024benchmarking}, i.e., we use a fully convolutional neural net. For the \gls{cstr}, we directly use the signals, concatenated into a $1400-$dimensional vector as the features. The data for these benchmarks is publicly available here\footnote{\url{https://www.kaggle.com/datasets/eddardd/tennessee-eastman-process-domain-adaptation}}, and here\footnote{\url{https://www.kaggle.com/datasets/eddardd/continuous-stirred-tank-reactor-domain-adaptation}}.

Note that, for each dataset, there are $n_{domains}(n_{domains}-1)$ adaptation tasks, except for \gls{cstr}, which is a multi-target benchmark (i.e., a single source, and 6 targets). As a result, we have $12 \times 3 + 6 \times 2 + 30 + 6 = 84$ adaptation tasks.

\newpage

\section{Detailed Results}

\begin{table}[ht]
    \resizebox{\linewidth}{!}{\begin{tabular}{ccccccccccc}
        \toprule
        Benchmark & Task & Baseline & OTDA$_{EMD}$ & OTDA$_{Sink}$ & OTDA$_{affine}$ & InfoOT$_{b}$ & InfoOT$_{c}$ & HOT-DA & GMM-OTDA$_{MAP}$ & GMM-OTDA$_{T}$\\
        \midrule
        \multirow{13}{*}{Caltech-Office} & $A \rightarrow D$ & 87.10 & 77.42 & 87.10 & 93.55 & 93.55 & 83.87 & \textbf{96.77} & 93.55 & 80.65\\
                                         & $A \rightarrow W$ & 91.53 & 93.22 & \textbf{96.61} & \textbf{96.61} & \textbf{96.61} & 94.92 & \textbf{96.61} & 93.22 & 91.53\\
                                         & $A \rightarrow C$ & 88.44 & \textbf{91.56} & 73.33 & 91.11 & 87.11 & 90.67 & 74.67 & 88.44 & 88.44\\
                                         & $D \rightarrow A$ & 88.54 & 90.10 & 93.23 & 92.19 & 91.15 & 92.71 & \textbf{96.88} & 96.35 & 95.83\\
                                         & $D \rightarrow W$ & 98.31 & 93.22 & 93.22 & 94.92 & \textbf{100.00} & 91.53 & 94.92 & \textbf{100.00} & 98.31\\
                                         & $D \rightarrow C$ & 75.56 & 67.11 & 17.33 & 71.56 & 68.44 & 75.11 & 51.11 & \textbf{86.67} & 71.56\\
                                         & $W \rightarrow A$ & 85.94 & 82.81 & 15.62 & 81.25 & 60.94 & 86.46 & 69.79 & \textbf{87.50} & 84.90\\
                                         & $W \rightarrow D$ & \textbf{100.00} & 93.55 & 93.55 & 96.77 & 90.32 & 93.55 & 90.32 & 96.77 & 93.55\\
                                         & $W \rightarrow C$ & 84.89 & 87.56 & 87.56 & \textbf{88.44} & 87.56 & 87.56 & 84.89 & 87.56 & 88.00\\
                                         & $C \rightarrow A$ & \textbf{98.44} & 96.88 & \textbf{98.44} & \textbf{98.44} & \textbf{98.44} & 96.88 & \textbf{98.44} & \textbf{98.44} & \textbf{98.44}\\
                                         & $C \rightarrow D$ & 93.55 & 87.10 & 90.32 & 93.55 & 87.10 & 87.10 & \textbf{100.0} & 93.55 & 80.65\\
                                         & $C \rightarrow W$ & 91.53 & 94.92 & 94.92 & 94.92 & 93.22 & 93.22 & 98.31 & \textbf{96.61} & 94.92\\
                                         & Avg.              & 90.32 & 87.95 & 78.44 & 91.11 & 87.87 & 89.46 & 87.72 & \textbf{93.22} & 88.90\\
        \midrule
        \multirow{13}{*}{ImageCLEF} & $B \rightarrow C$  & 92.50 & 95.00 & 95.00 & 94.17 & \textbf{97.50} & 96.67 & 96.67 & 95.00 & 94.17\\
                                    & $B \rightarrow I$  & 90.00 & 89.17 & 89.17 & 91.67 & 94.17 & 91.67 & \textbf{95.00} & \textbf{95.00} & 93.33\\
                                    & $B \rightarrow P$  & 68.33 & 69.17 & 70.00 & 71.67 & 73.33 & \textbf{75.83} & 74.17 & 72.50 & 74.17\\
                                    & $C \rightarrow B$  & 65.00 & 65.83 & 65.83 & 65.00 & 51.67 & 62.50 & 62.50 & 65.00 & \textbf{66.67}\\
                                    & $C \rightarrow I$  & 89.17 & 96.67 & 96.67 & 95.00 & 92.50 & \textbf{97.50} & 95.83 & 94.17 & 96.67\\
                                    & $C \rightarrow P$  & 71.67 & 74.17 & 73.33 & 71.67 & 75.00 & \textbf{75.83} & 72.50 & 70.83 & \textbf{75.83}\\
                                    & $I \rightarrow B$  & 68.33 & \textbf{70.00} & 68.33 & \textbf{70.00} & 65.00 & 66.67 & 61.67 & 67.50 & \textbf{70.00}\\
                                    & $I \rightarrow C$  & 93.33 & 95.83 & 95.83 & 95.83 & 95.83 & \textbf{96.67} & 95.83 & 95.00 & 95.83\\
                                    & $I \rightarrow P$  & 71.67 & 74.17 & \textbf{75.00} & 73.33 & 73.33 & 71.67 & 72.50 & 73.33 & \textbf{75.00}\\
                                    & $P \rightarrow B$  & 67.50 & \textbf{69.17} & 66.67 & 68.33 & 57.50 & 65.83 & 62.50 & 62.50 & 64.17\\
                                    & $P \rightarrow C$  & 95.00 & 95.83 & 95.83 & 95.00 & \textbf{96.67} & \textbf{96.67} & \textbf{96.67} & 95.00 & 95.00\\
                                    & $P \rightarrow I$  & 90.83 & 92.50 & 92.50 & 90.83 & \textbf{95.83} & \textbf{95.83} & 95.00 & 94.17 & 95.00\\
                                    & Avg.               & 80.28 & 82.29 & 82.01 & 81.88 & 80.69 & 82.78 & 81.74 & 81.67 & \textbf{82.99}\\
       \midrule
        \multirow{7}{*}{Office 31} & $A \rightarrow D$ & 66.07 & 68.75 & 69.64 & 69.64 & 75.89 & \textbf{76.79} & 72.32 & 69.64 & 72.32\\
                                   & $A \rightarrow W$ & 76.02 & 74.27 & 80.12 & 80.12 & 79.53 & 79.53 & 73.68 & 76.61 & \textbf{80.70}\\
                                   & $D \rightarrow A$ & 65.68 & 65.85 & 67.77 & 66.90 & 67.60 & 66.20 & 61.15 & 68.29 & \textbf{73.52}\\
                                   & $D \rightarrow W$ & 94.15 & 95.32 & 98.25 & 98.25 & 95.91 & 97.08 & 84.80 & \textbf{98.83} & 95.32\\
                                   & $W \rightarrow A$ & 63.41 & 66.90 & 67.42 & 65.51 & \textbf{67.60} & \textbf{67.60} & 61.67 & 66.38 & 65.68\\
                                   & $W \rightarrow D$ & \textbf{96.43} & 90.18 & 92.86 & 95.54 & 87.50 & 91.96 & 81.25 & 91.96 & 91.07\\
                                   & Avg.              & 76.96 & 76.88 & 79.34 & 79.32 & 79.00 & \textbf{79.86} & 72.48 & 78.62 & 79.77\\
        \midrule
        \multirow{13}{*}{Office-Home} & $Ar \rightarrow Cl$  & 55.10 & 54.98 & 54.87 & 56.24 & 17.41 & 53.95 & 47.88 & 53.95 & 57.96\\
                                      & $Ar \rightarrow Pr$  & 70.95 & 68.69 & 71.96 & 71.96 & 30.97 & 70.27 & 67.23 & 74.89 & 74.10\\
                                      & $Ar \rightarrow Rw$  & 79.68 & 79.68 & 80.83 & 80.71 & 40.53 & 80.25 & 76.00 & 77.96 & 82.43\\
                                      & $Cl \rightarrow Ar$  & 63.51 & 60.62 & 63.09 & 62.68 & 31.34 & 62.68 & 53.81 & 59.79 & 64.33\\
                                      & $Cl \rightarrow Pr$  & 69.26 & 66.89 & 68.81 & 70.72 & 41.78 & 68.92 & 63.51 & 70.05 & 71.73\\
                                      & $Cl \rightarrow Rw$  & 72.68 & 69.92 & 71.18 & 72.33 & 38.81 & 71.18 & 67.97 & 68.66 & 74.63\\
                                      & $Pr \rightarrow Ar$  & 66.80 & 62.47 & 64.12 & 66.39 & 32.16 & 64.33 & 55.88 & 57.73 & 62.89\\
                                      & $Pr \rightarrow Cl$  & 36.88 & 38.83 & 25.32 & 38.83 & 8.59  & 30.93 & 23.71 & 30.70 & 31.62\\
                                      & $Pr \rightarrow Rw$  & 78.76 & 77.84 & 79.22 & 79.22 & 47.99 & 78.30 & 71.64 & 73.59 & 80.94\\
                                      & $Rw \rightarrow Ar$  & 72.99 & 71.96 & 72.37 & 73.81 & 51.13 & 70.72 & 62.47 & 66.39 & 69.69\\
                                      & $Rw \rightarrow Cl$  & 53.15 & 57.85 & 57.39 & 56.93 & 37.00 & 56.59 & 47.65 & 50.63 & 56.36\\
                                      & $Rw \rightarrow Pr$  & 82.32 & 80.86 & 81.87 & 82.21 & 64.86 & 81.31 & 72.30 & 80.41 & 82.09\\
                                      & Avg.                 & 66.84 & 65.88 & 65.92 & 67.67 & 36.88 & 65.79 & 59.17 & 63.73 & 67.40\\
        \bottomrule
    \end{tabular}}
    \caption{Single-source domain adaptation results. We compare 8 methods over 5 benchmarks, with a total of 42 adaptation tasks.}
    \label{tab:single_source_results}
\end{table}

\end{document}